\documentclass[10pt,letterpaper]{article}

\ifdefined\localcompile\else
  \pdfoutput=1
\fi

\usepackage[margin=1in]{geometry}
\usepackage{setspace}
\usepackage{fancyhdr}

\usepackage[cmex10]{amsmath}
\usepackage{amssymb}
\usepackage{nicefrac}

\usepackage{graphicx}
\usepackage[hyperref,x11names,cmyk,table,pdftex]{xcolor}
\usepackage[hypcap=true]{caption}
\usepackage[hypcap=true]{subcaption}

\usepackage{booktabs}
\usepackage{multirow}

\usepackage{enumerate}
\usepackage{paralist}

\usepackage{times}
\usepackage[resetfonts]{cmap}
\usepackage[T1]{fontenc}
\usepackage[utf8]{inputenc}

\ifdefined\draftmode
  \usepackage[layout={inline},author=]{fixme}
\else
  \usepackage[final,layout={inline},author=]{fixme}
\fi

\usepackage[breaklinks=true,colorlinks,bookmarks=true]{hyperref}
\usepackage{cite}
\usepackage{cleveref}

\title{ 
  Hyperspectral Image Classification and Clutter Detection via
  Multiple Structural Embeddings and Dimension Reductions
 }

\makeatletter
\let\titletext\@title
\makeatother

\ifdefined\draftmode    

  \author{
    Alexandros-Stavros~Iliopoulos       \and
    Tiancheng~Liu                       \and
    Xiaobai~Sun
  }

\else

  \author{
    Alexandros-Stavros~Iliopoulos%
    \thanks{A.S.\ Iliopoulos and X.\ Sun are with the department of
      Computer Science, Duke University, Durham, NC 27708, USA.}%
    \and
    Tiancheng~Liu%
    \thanks{T.\ Liu is with the department of Electrical and Computer
      Engineering, Duke University, Durham, NC 27708, USA.}%
    \and
    Xiaobai~Sun%
    \samethanks[1]%
  }

\fi     

\newcommand{\datestring}{February 24, 2015}

\ifdefined\draftmode            

  \date{\datestring}

\else                           

  \date{}

\fi

\newcommand{\pdfauthors}{A.S.\ Iliopoulos, T.\ Liu, X.\ Sun}

\hypersetup{%
  pdffitwindow=false,%
  pdfstartview={FitH},%
  colorlinks,%
  pdfauthor={\pdfauthors},%
  pdftitle={\titletext},%
  pagebackref=false,%
  citecolor=PaleGreen4,%
  filecolor=DarkOrchid4,%
  linkcolor=OrangeRed4,%
  urlcolor=black
}

\graphicspath{%
  {./}%
  {../../../figs/pdf/}%
}

\pdfpageattr{/Group <</S /Transparency /I true /CS /DeviceRGB>>}

\let\leftorig\left
\let\rightorig\right
\renewcommand{\left}{\mathopen{}\mathclose\bgroup\leftorig}
\renewcommand{\right}{\aftergroup\egroup\rightorig}

\newcommand{\samethanks}[1][\value{footnote}]{\footnotemark[#1]}

\fxsetup{inline}
\fxusetheme{color}
\newcommand{\note}[1]{\fxerror{\ [#1]}}

\ifdefined\draftmode
  \setkeys{Gin}{draft=false}
\fi

\ifdefined\draftmode
  \hypersetup{final}
\fi

\newlength{\vshiftmrow}
\newcommand{\hshiftacc}{\hspace*{4.5pt}}

\newcommand{\fignocbar}{0.22\linewidth}
\newcommand{\figcbar}{0.26\linewidth}
\newlength{\heightIndianPines}
\newlength{\heightPaviaU}
\newcommand{\shiftcbarIndianPines}{\hspace*{1.95em}}
\newcommand{\shiftcbarPaviaU}{\hspace*{3.4em}}
\newlength{\heightIndianPinesBig}
\newlength{\heightPaviaUBig}
\newcommand{\mbs}[1]{\mathbf{#1}}
\newcommand{\longvec}[4][]{%
  \left[#2_1#1 \; \cdots \; #2_{#3}#1 \; \cdots \; #2_{#4}#1 \right]}

\newcommand{\normFro}[1]{\left\|#1\right\|_F}
\newcommand{\normTwo}[1]{\left\|#1\right\|_2}

\newcommand{\threshold}{\tau}
\newcommand{\vecZero}{\mbs{0}}
\newcommand{\matel}[2]{\left[#1_{#2}\right]}

\newcommand{\dataAmb}{x}
\newcommand{\dataMan}{y}

\newcommand{\vecAmb}{\mbs{\dataAmb}}

\newcommand{\matAmb}{\mbs{\MakeUppercase{\dataAmb}}}
\newcommand{\matMan}{\mbs{\MakeUppercase{\dataMan}}}

\newcommand{\basisFeat}{\phi}
\newcommand{\setBasisFeat}{\Phi}        
\newcommand{\embfeat}[1]{\hat{#1}}

\newcommand{\dimAmb}{D}
\newcommand{\dimMan}{d}

\newcommand{\patchSize}{p}

\newcommand{\localNhood}{\mathcal{N}}
\newcommand{\weightLLE}{w}

\newcommand{\matWeightLLE}{\mbs{\MakeUppercase{\weightLLE}}}
\newcommand{\matIdentity}{\mbs{I}}
\newcommand{\matEmbLLE}{(\matIdentity - \matWeightLLE)}
\newcommand{\vecConst}{\mbs{e}}
\newcommand{\sizeNhood}{k}

\newcommand{\countLabel}{n}
\newcommand{\freqLabel}{f}
\newcommand{\numLabels}{L}
\newcommand{\numTests}{T}
\newcommand{\ambiguity}{H}

\newcommand{\thresClutter}{\threshold_{\text{clt}}}

\newcommand{\numPixels}{N}
\newcommand{\numBases}{M}

\newcommand{\idxFeatEmb}{i}
\newcommand{\idxPix}{j}
\newcommand{\idxNhood}{i}
\newcommand{\idxLabel}{\ell}

\newcommand{\knn}{$k$NN}

\begin{document}

\ifdefined\draftmode
  \doublespacing
\fi

%
\pdfbookmark[1]{Title}{sec:Title}
\maketitle

%
\phantomsection
\label{sec:abstract}
\addcontentsline{toc}{section}{Abstract}

\begin{abstract}

We present a new and effective approach for Hyperspectral Image (HSI)
classification and clutter detection, overcoming a few long-standing
challenges presented by HSI data characteristics.
Residing in a high-dimensional spectral attribute space, HSI data samples
are known to be strongly correlated in their spectral signatures, exhibit
nonlinear structure due to several physical laws, and contain uncertainty
and noise from multiple sources.
In the presented approach, we generate an adaptive, structurally enriched
representation environment, and employ the locally linear embedding (LLE)
in it.
There are two structure layers external to LLE.
One is feature space embedding: the HSI data attributes are embedded into a
discriminatory feature space where spatio-spectral coherence and
distinctive structures are distilled and exploited to mitigate various
difficulties encountered in the native hyperspectral attribute space.
The other structure layer encloses the ranges of algorithmic parameters for
LLE and feature embedding, and supports a multiplexing and integrating
scheme for contending with multi-source uncertainty.
Experiments on two commonly used HSI datasets with a small number of
learning samples have rendered remarkably high-accuracy classification
results, as well as distinctive maps of detected clutter regions.



\end{abstract}

%
\section{Introduction}
\label{sec:introduction}


We are concerned in this paper with analysis of hyperspectral imaging (HSI)
data; in particular, we address the task of high-accuracy multi-class
labeling, as well as clutter detection as a necessary complement.

Enabled by advanced sensing systems, such as the NASA/JPL
AVIRIS~\cite{green_imaging_1998}, NASA
Hyperion~\cite{pearlman_overview_2001}, and DLR
ROSIS~\cite{holzwarth_hysens_2003} sensors, hyperspectral imaging, also
known as imaging spectroscopy, pertains to the acquisition of
high-resolution spectral information over a broad range, providing
substantially richer data than multi-spectral or color imaging. HSI
combines spectral with spatial information, as samples are collected over
large areas, at increasingly fine spatial resolution.
With its rich information provision and non-invasive nature, HSI has become
an invaluable tool for detection, identification, and classification of
materials and objects with complex compositions. Relevant application
fields include material science, agriculture, environmental and urban
monitoring, resource discovery and monitoring, food safety and security,
and medicine~\cite{bioucas-dias_hyperspectral_2012, mohan_spatially_2007,
  huang_recent_2014}.
As sensing technologies continue to advance, HSI is providing larger
collections of data to facilitate and enable scientific and engineering
inquiries that were previously unfeasible. At the same time, it challenges
many existing data analysis methods to render high-quality results
commensurate with the richness of available information in HSI data.

Among the key challenging factors for HSI data analysis are:
the curse of dimensionality of the spectral feature space, which hampers
class discrimination (Hughes effect~\cite{hughes_mean_1968}) and
exacerbates the computational complexity of the analysis process;
strong and nonlinear spatio-spectral correlations and mixing across
spectral bands, as well as cross-mixing between spatial pixels and spectral
bands~\cite{bioucas-dias_hyperspectral_2013};
and multiple sources of noise and uncertainty with regard to the imaged
scene and acquisition process~\cite{bachmann_exploiting_2005}.

A host of data analysis approaches have been investigated for use in HSI
classification~\cite{bioucas-dias_hyperspectral_2013}.
We may roughly categorize them according to the feature space where
classification takes place and whether or not the corresponding models are
linear. For example, band selection and linear combination techniques for
classification~\cite{jimenez-rodriguez_unsupervised_2007,
  kumar_best-bases_2001, tadjudin_covariance_1998} reduce the
dimensionality of the spectral attribute space based on linear signal and
image models.
Kernel-based classifiers, such as SVMs~\cite{li_multiple_2015,
  camps-valls_kernel-based_2005}, respect nonlinearity, applying a
nonlinear transform to the data attributes and embedding them in a
high-dimensional classification space. While such methods can be effective
with certain data, they can be sensitive to the chosen embedding kernel and
the number or distribution of available training samples.
A different approach is that of manifold learning
methods~\cite{roweis_nonlinear_2000, tenenbaum_global_2000,
  ni_classification_2015, mohan_spatially_2007, bachmann_exploiting_2005},
where high-dimensional embedding of the data samples is followed by
dimensionality reduction. Such methods assume that a (principal) manifold
structure underlies the collected data samples; subsequent analyses are
then based on the isometric principles associated with manifolds. Another
important assumption is that the features lie in a well-defined metric
space; manifold learning methods are sensitive to the choice of metric for
neighborhood definition, as well as to the density and distribution of data
samples.
Indeed, a naive application of such an approach to HSI data may suffer from
the high correlation and various uncertainty sources in the hyperspectral
attribute space.
It should be noted, additionally, that some algorithms incur too high
a computational cost for them to be practical for HSI data analysis,
even more so as the spatial coverage and resolution of hyperspectral
sensors is increasing.

We address here the aforementioned standing issues in HSI
classification:%
\begin{inparaenum}[(i)]
\item nonlinear correlation and irregular singularities,
\item multiple-source uncertainties with respect to the HSI data structure,
  and 
\item high data dimensionality.
\end{inparaenum}
The first problem is in part responsible for an existing gap between HSI
data collection and analysis: while spectral and spatial information is
coupled in HSI scenes, it is typically processed in a decoupled manner.
From an alternative perspective, the strong correlation in HSI data can be
exploited to help overcome the other two challenges. In our approach, we
start with exploring and utilizing the spatial and spectral coherence of
HSI data in tandem.
There are various methods that attempt to incorporate spatial
coherence~\cite{fauvel_advances_2013, tarabalka_multiple_2010,
  kang_spectral-spatial_2014, mohan_spatially_2007} in the analysis
process; these approaches can be seen as special or extreme cases in the
framework we introduce in this paper.

There are three key components in our framework for HSI classification and
clutter detection:%
\begin{inparaenum}[(i)]
\item The Locally Linear Embedding (LLE) method of Roweis and
  Saul~\cite{roweis_nonlinear_2000} provides the basic computational
  procedure for deriving a manifold representation of the data; we review
  LLE in \cref{sec:lle-review} and comment on our interpretation, our
  rationale for its selection, and its specific form within our method.
\item Prior to the LLE computations, we embed the HSI samples to a
  structural feature space using efficient, local filters to highlight
  their spatio-spectral structure, thus exposing potential discriminatory
  singularities and contending with noise in the data, while avoiding
  de-correlation; we describe the feature embedding concept and its
  connection to the LLE processing in \cref{sec:feature-embeddings}.
\item We consider an ensemble of structural embeddings and representations,
  defined by multiple parameter instances for the other two components, to
  counteract the effect of multiple uncertainties; we describe in
  \cref{sec:ensemble} the relevant ensemble parameters, as well as our
  scheme for multiplexing and integrating the results over all instances.
\end{inparaenum}

Experimental results with our approach are presented in
\cref{sec:experiments}. They demonstrate evidently high-accuracy
classification and clutter detection. Indeed, the estimated clutter
maps we extract appear to be the first of their kind in the context of
HSI classification. Clutter areas shape boundaries and delineate
coherent, labeled regions; they may also contain objects of interest
or new classes to be analyzed, and may be of higher value to various
data analysis applications.
We consider clutter maps, such as the ones presented in this paper, as
critical information that complements classification in the traditional
sense. The joint provision of classification and clutter detection
estimates serves to make HSI data analysis independent of artificial or
impractical conditions, and impacts the rendering of higher quality,
interpretable analysis results.



%
\section{The LLE method for classification}
\label{sec:lle-review} 


The core processing module for HSI structure encoding and classification in
our approach is the Locally Linear Embedding (LLE) method of Roweis and
Saul~\cite{roweis_nonlinear_2000}. The basic assumption behind it is that a
set of data samples in a high-dimensional space of observable attributes is
distributed over an underlying low-dimensional manifold; LLE may then be
used to map the data samples to the principal manifold coordinate space, or
parameter space. This assumption conforms well to HSI data, owing to their
non-linear, correlated structure, as per the physical laws of radiative
transfer and sensor properties and
calibration~\cite{bachmann_exploiting_2005,
  bioucas-dias_hyperspectral_2013}, whereas direct use of linear dimension
reduction models is ill-suited for HSI data analysis.

LLE has rendered surprisingly good results in classification or clustering
of synthetic data samples on low-dimensional manifolds (e.g.\ Swiss roll)
and certain image data (such as handwritten digits and facial pose or
illumination)~\cite{roweis_nonlinear_2000,chang_robust_2006}. Several
theoretical interpretations and algorithmic extensions have been proposed
for LLE~\cite{donoho_hessian_2003, bengio_out--sample_2003}, and it is
increasingly applied to domain-specific data analysis tasks. HSI
classification ranks among such tasks~\cite{mohan_spatially_2007,
  kim_hyperspectral_2003}, albeit scarcely.

In this work, we adopt LLE as a core procedure for HSI classification due
to three of its remarkable properties:%
\begin{inparaenum}[(i)]
\item the natural connection between a globally connected embedding of
  local geometric structures and sparse coding;
\item the translation invariance of local geometry encoding and its
  preservation by dimensionality reduction; and
\item the strikingly simple and computationally efficient algorithmic
  structure.
\end{inparaenum}
We briefly describe the LLE processing steps and remark on certain aspects
based on our interpretation.

Let $\matAmb = \longvec{\vecAmb}{\idxPix}{\numPixels}$, where
$\vecAmb_\idxPix \in \mathbb{R}^\dimAmb$, be a set of $\numPixels$
samples in a $\dimAmb$-dimensional feature space. First, a set of
neighboring samples, denoted by $\localNhood_\idxPix$, is located for
every sample, $\vecAmb_\idxPix$. We employ the $\sizeNhood$-nearest
neighbors (\knn) scheme because of its relative insensitivity to
sample density; our measure for neighborhood definition is based on
angular (cosine) similarity.

The local geometry around each sample point, $\vecAmb_\idxPix$, is then
encoded by a vector of local coefficients (weights). These coefficients
place $\vecAmb_\idxPix$ at the neighborhood barycenter and the
corresponding vector is numerically orthogonal to the tangent plane spanned
by its neighbors about the center. Specifically, the local weights,
$\weightLLE_{\idxNhood\idxPix}$, are determined by the following local
least squared problem, subject to the affine combination condition:
\begin{align}
  \label{eq:lle-local-weights}
  \min_{\{\weightLLE_{\idxNhood\idxPix}\}} 
  \normTwo{
  \vecAmb_\idxPix - 
  \sum_{\idxNhood \in \localNhood_\idxPix}
  \weightLLE_{\idxNhood\idxPix} \vecAmb_\idxNhood
  }^2 ,
  &&
  \text{s.t.} \;\;
  \sum_{\idxNhood \in \localNhood_\idxPix}
  \weightLLE_{\idxNhood\idxPix} = 1 ,
\end{align}
for all $\idxPix \in \{1,\ldots,\numPixels\}$. The affine combination not
only makes the sample point the neighborhood barycenter, but also means
that the local encoding is translation invariant.

\Cref{eq:lle-local-weights} may be rewritten in matrix form as
\begin{align}
  \label{eq:lle-weight-aggregation}
  \min_\matWeightLLE 
  \normFro{ \matAmb \matEmbLLE }^2 \; ,
  &&
  \text{s.t.} \;\;
  \vecConst^\top \matEmbLLE = \vecZero ,
\end{align}
where $\normFro{\cdot}$ is the Frobenius norm, $\matIdentity$ is the
identity matrix, $\vecConst$ is the constant-1 vector, and $\matWeightLLE$
is an $\numPixels \times \numPixels$ matrix,
$\matWeightLLE = \matel{\weightLLE}{\idxNhood\idxPix}$.

Once $\matWeightLLE$ is computed, the left singular vectors,
$\matMan$, corresponding to the ($\dimMan + 1$) smallest singular
values of $\matEmbLLE$ are obtained:
\begin{align}
  \label{eq:lle-dimension-reduction}
  \min_\matMan
  \normFro{ \matMan \matEmbLLE }^2 \; ,
  &&
  \text{s.t.} \;\;
  \matMan \matMan^\top = \matIdentity_{\dimMan+1} ,
\end{align}
where $\dimMan < \dimAmb$ is the reduced dimensionality. The
low-dimensional representation, $\matMan$, of the data samples preserves
local geometry and global connectivity as encoded in $\matEmbLLE$.

Finally, a classifier is employed to label the data in the low-dimensional
manifold parameter space. We use a simple nearest-neighbor classifier to
investigate the efficacy of the embedding and dimension reduction process
with respect to classification.

A few additional remarks: The sparsity pattern of the weight matrix,
$\matWeightLLE$, is determined by the \knn\ search in the first step, while
the corresponding numerical values of $\matWeightLLE$ are determined via
\cref{eq:lle-local-weights} in a local, column-wise independent
fashion. More importantly, $\matWeightLLE$, as per
\cref{eq:lle-weight-aggregation}, encodes the global interconnection of
local hyperplanes via the transitive property of neighborhood connections,
without entailing the explicit, computationally expensive calculation of
all pairwise shortest connection paths. The $\matWeightLLE$ matrix can also
be seen as a simple kernel-based embedding.
%
%
The low-dimensional space spanned by $\matMan$ includes
constant-valued vectors, corresponding to the zero singular value,
whose geometric multiplicity may be greater than $1$. The
discriminatory information lies in the $\dimMan$-dimensional subspace
that is orthogonal to the constant vector, $\vecConst$.
%



%
\section{Feature space embedding} 
\label{sec:feature-embeddings} 


HSI data samples are known to be strongly correlated in their spectral
signatures~\cite{bioucas-dias_hyperspectral_2012, kumar_best-bases_2001,
  bachmann_exploiting_2005}. Strong correlation between features
complicates the choice of a discriminatory distance or similarity metric,
particularly so in a high-dimensional setting. Furthermore, nonlinearity
and high dimensionality render de-correlation attempts
ineffective. Increasing the learning sample density is impractical and may
yield limited improvements; learning from sparse reference sample subsets
is desired, instead.

We take a novel approach, namely structural feature embedding, to alleviate
these fundamental issues. We explore the spatio-spectral coherence
structure of HSI data, and embed the spectral attribute space in a
structure-rich space, where data-specific features may be made more
salient. Then, a conventional distance metric in the embedding feature
space may be seen as an \emph{ad hoc} discriminatory one in the original
attribute space. Moreover, the computational complexity for structural
feature embedding scales linearly with the dataset size, which is much more
efficient than that of even linear de-correlation.

Specifically, we explore spatial and spectral coherence by using a bank of
filters.  Formally, the filters define a set of basis (or transform)
functions, $\setBasisFeat$, such that the embedded data become
\begin{equation}
  \label{eq:feature-space-embedding}
  \embfeat{\matAmb} = 
  \setBasisFeat( \matAmb ) =
  \longvec[(\matAmb)]{\basisFeat}{\idxFeatEmb}{\numBases}^\top
  ,
\end{equation}
where each basis, $\basisFeat_\idxFeatEmb$, is local with respect to the
spatial and/or spectral domain of the HSI dataset, $\matAmb$.
Thus, the embedded feature space may be efficiently computed, removing
certain noise components while preserving the underlying manifold
structure. The distance or similarity between any two samples is then
measured in the embedded feature space.

Feature transformation and embedding directly impact the metric for
neighborhood definition and subsequent encoding of local geometry. A
closely related notion with respect to the spatial properties of the HSI is
the spatially coherent distance function introduced by Mohan \emph{et
  al.}~\cite{mohan_spatially_2007}, where it is proposed that distance
calculations be performed using all features in a local, ordered patch
around each pixel. Here, we introduce the notion of feature embedding as a
basic mechanism for effecting a data-specific geometric metric by means of
a conventional metric, thus circumventing the explicit definition of new,
complicated metrics.
Note, for example, that employing the patch-based spatially coherent
distance of Mohan \emph{et al.}\ is equivalent to applying a box filter to
each HSI band prior to distance calculations---except that the latter is
insensitive to the particular ordering of pixels within the patch, making
similarity discovery more robust with respect to local composition
variations and object boundaries.

In general, the feature transform basis functions, or simply filters, can
be divided into two groups: generic ones that may be useful to any HSI
analysis task, and data- or analysis-specific filters, depending on one's
objective.
The filters can be also grouped according to their geometric and
statistical features. We consider two particular types of spectral
filters: differential and integral. Differential filters elucidate
local characteristics of the spectral signature of each sample, and
generally down-weigh spurious similarity contributions induced by
correlation between consecutive spectral bands. Integral filters, on
the other hand, may be used to extract statistical, noise-insensitive
properties of spectral signatures.

This embedding mechanism allows us to probe the HSI data at different
scales, depending on the support and order of the spatial or spectral
filters; hence, the hyperspectral data are embedded in a feature space
that captures their structure at the relevant scale.
In the experiments carried out in this paper, we use spatial box filtering,
and extend the spectral features with their numerical gradient and first
two statistical moments (mean and standard deviation).

A few remarks are in order on the computation of local neighborhoods.
Obtaining the local neighborhoods, $\localNhood$, which directly affect the
estimated manifold structure and parameters, amounts to computation of all
$k$-nearest neighbors sets among the hyperspectral samples. This starts to
become problematic as the size of the HSI increases, due to the high
computational cost of \knn\ searching in the high-dimensional embedded (or
original) feature space. Based on the spatial coherence of HSIs---and given
that the size of each local neighborhood should be relatively small for the
approximately linear structure assumption to hold in its vicinity---we
circumvent this issue by bounding the search for spectral neighbors within
an ample spatial window centered around each pixel.%
\note{Owing to the sparsity of the LLE matrix, we may still generate the
  full, connected system of local hyperplanes and the consequent
  low-dimensional representation of the dataset, without needing to resort
  to tiling and local-coordinate
  transformations~\cite{bachmann_exploiting_2005}.}




%
\begin{table*}[t]
  \caption{HSI dataset summary.}
  \label{tab:datasets}
  \centering
  \small
  \begin{tabular}{%
    c 
    c 
    r @{--} l
    c 
    c 
    r @{$\times$} l
    c 
    c 
    c
    }
    \toprule
      \multirow{2}{*}[\vshiftmrow]{\textbf{Dataset}}
    & \multirow{2}{*}[\vshiftmrow]{\textbf{Sensor}}
    & \multicolumn{4}{c}{\textbf{Spectral domain} ($nm$)} 
    & \multicolumn{3}{c}{\textbf{Spatial domain} $\left(m^2\right)$ } 
    & \multirow{2}{*}[\vshiftmrow]{\textbf{\#classes}} 
    & \multirow{2}{*}[\vshiftmrow]{\parbox{5.6em}{%
      \centering\bfseries Labeled area coverage}}
    \\
    \cmidrule(lr){3-6}
    \cmidrule(lr){7-9}
    &
    & \multicolumn{2}{c}{\textbf{range}}
    & \textbf{resolution}
    & \textbf{\#bands}
    & \multicolumn{2}{c}{\textbf{\#pixels}}
    & \textbf{resolution}
    &
    &
    \\
    \midrule
      Indian Pines
    & AVIRIS~\cite{green_imaging_1998}
    & 410 & 2450
    & 10
    & 220
    & 145 & 145
    & 200
    & 16
    & 49.4\%
    \\
      Univ.\ of Pavia
    & ROSIS~\cite{holzwarth_hysens_2003}
    & 430 & 860
    & 4
    & 103
    & 610 & 340
    & 1.7
    & 9
    & 20.6\%
    \\
    \bottomrule
  \end{tabular}
\end{table*}




%
\section{Structural algorithm ensemble}
\label{sec:ensemble}


As has already been mentioned, there are multiple sources that introduce
variations and uncertainty to the underlying HSI manifold structure. To
name a few, such variations may stem from scattering, atmospheric
conditions, spectral mixing of material constituents,
etc~\cite{bachmann_exploiting_2005,
  bioucas-dias_hyperspectral_2013}. Another related issue is that HSI
samples pertaining to different compounds may be distributed
inhomogeneously along the manifold surface.
The introduction of uncertainty from a diverse set of sources to the
observed HSI attributes means that the sample manifold will tend to exhibit
multi-scale structure.
These considerations motivate us to probe the HSI data at different scales
in order to uncover the underlying structure.

The derived HSI representation depends on several parameters in all stages
of the embedding and dimension reduction procedure, each capturing
different properties of the HSI manifold:%
\begin{inparaenum}[(i)]
\item the choice of spatial and spectral filter parameters determines the
  type and scale of features that define similarity between samples;
\item the size of local neighborhoods, relative to the sample distribution
  density around each sample, defines the coarseness and connectivity of
  the manifold encoding in the embedded feature space; and
\item the dimensionality of the parametrized manifold representation
  affects the type of manifold features that are used for classification.
\end{inparaenum}

We define a relevant search space for the set of these algorithmic
parameters and obtain an ensemble of structural embeddings and
low-dimensional manifold representations of the HSI data. For all HSI samples,
we find the label of their nearest reference sample in each representation
instance. This set of proximity labels is then used to obtain the
classification results, together with a clutter map estimate.




\subsection{Classification entropy and clutter estimation}
\label{sec:multiplexing-ambiguity-classification}


Hyperspectral image scene classification methods typically assign each
pixel in the imaged scene to one of the classes for which labeled reference
samples in the scene (also known as ground truth) are
available. Oftentimes, however, a large portion of the HSI may be comprised
of pixels that belong to none of the labeled classes; these pixels
constitute \emph{clutter} with respect to the specified label-set. Clutter
pixels are likely diverse in terms of their spectral features, and cannot
generally be considered to correspond to a single, new class.
A related but somewhat different approach is taken in the context of
anomaly detection. There, identification of the ``clutter'' (anomalous)
region typically depends on the collection and utilization of statistical
properties of relevant scenes, obtained from a large set of learning
examples~\cite{chandola_anomaly_2009}.
Here, we do not require additional data beyond those in a single HSI data
cube, and restrict the reference/learning samples, used for classification,
to a sparse subset of available data samples.

For classification and clutter detection, we first obtain a classification
entropy score for every non-reference pixel, as follows. Each non-reference
pixel is matched to its nearest (in the low-dimensional classification
space) reference pixel, for all instances, or trials, that make up our
ensemble.
Hence, given a total of $\numTests$ trials, each pixel is associated with a
vector of $\numTests$ proximity labels. This vector is converted to a
frequency vector of length $L$, where $L$ is the number of labeled
classes. Let $\countLabel_{\idxPix\idxLabel}$ be the count of the
$\idxLabel$-th label, $\idxLabel \in \{1, \ldots, \numLabels\}$, in
proximity-label vector of the $\idxPix$-th pixel, and
$\freqLabel_{\idxPix\idxLabel} =
\frac{\countLabel_{\idxPix\idxLabel}}{\numTests}$
be the corresponding relative frequency.
Taking an information-theoretic approach, we define the classification
entropy for the $\idxPix$-th pixel as
\begin{equation}
  \label{eq:ambiguity-entropy-def}
  \ambiguity_\idxPix =
  -\sum_{\idxLabel = 1}^{\numLabels} {
    \freqLabel_{\idxPix\idxLabel} 
    \log_\numLabels (\freqLabel_{\idxPix\idxLabel})
  } .
\end{equation}
The classification entropy score $\ambiguity_\idxPix$ lies in $[0,1]$. At
one extreme ($\ambiguity_\idxPix = 0$), the labeling frequency vector of
the $\idxPix$-th pixel has only one non-zero element, meaning that all of
its proximity labels are the same. At the other extreme
($\ambiguity_\idxPix = 1$), the frequency vector is constant, meaning that
all proximity labels for the pixel are equally frequent among the
$\numTests$ instances or trials.
%
%
Empirically, $\ambiguity_\idxPix$ measures the classification ambiguity of
the $\idxPix$-th pixel. A pixel with a high classification entropy score is
most likely a clutter pixel, whereas a pixel with a low score is likely to
belong to one of the available classes. The $\ambiguity_\idxPix$ scores for
all pixels can be displayed as a grayscale image, providing an
classification entropy map for a given experimental ensemble---see
\cref{sec:exp-results}.

Using a threshold, $\thresClutter$, we make use of the classification
entropy map to split the HSI scene into two complementary parts: clutter regions
($\ambiguity_\idxPix \ge \thresClutter$), where no label is given
to the corresponding pixels, and
labeled regions ($\ambiguity_\idxPix < \thresClutter$), where each pixel is
matched to the available classes.
While a diverse set of methods has been proposed for
combining results in multiple classifier
systems~\cite{tin_kam_ho_decision_1994, briem_multiple_2002,
  tzeng_remote_2006, wozniak_survey_2014}, most rely on the availability of
enough training data or knowledge of certain statistical properties of the
dataset and/or classifiers, which may not be the case in many practical
applications. Here, we assign each pixel to the most frequently returned
class for it among the set of results for each classifier instance. This
simple rule provides us with a baseline regarding the performance of our
methodology; moreover, it does not entail additional assumptions or
abundance of labeled data, and we have found it to generally improve upon
any single classifier instance throughout our experiments.




%
\begin{figure*}[t]
  \centering
  %
  \begin{subfigure}[b]{\fignocbar}
    \centering
    \includegraphics[height=0.985\heightIndianPines]{%
      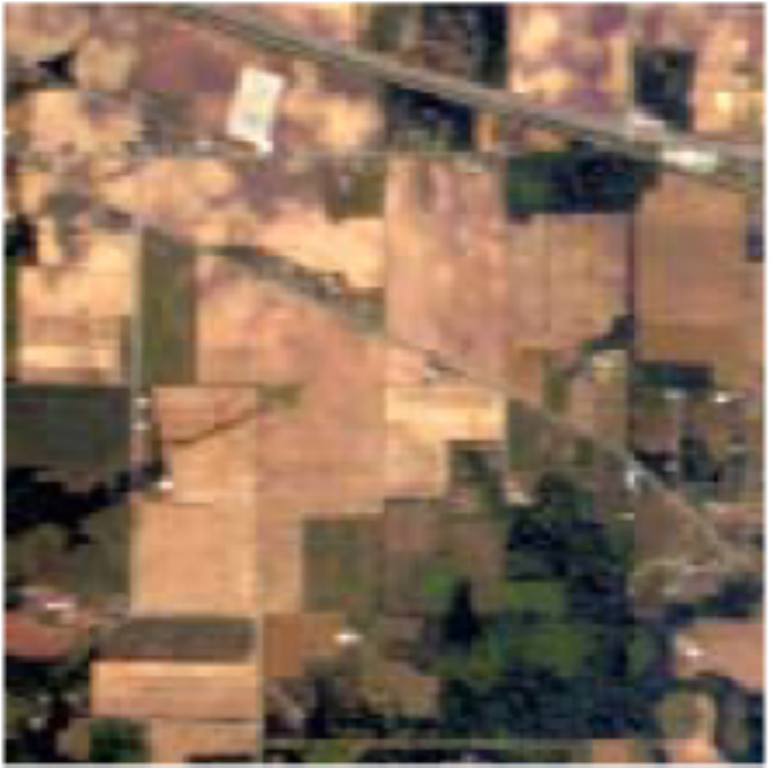}%
    \caption{ }
    \label{fig:IndianPines_traditional_RGB}
  \end{subfigure}
  \hfill\hfill
  %
  \begin{subfigure}[b]{\fignocbar}
    \centering
    \includegraphics[height=\heightIndianPines]{%
      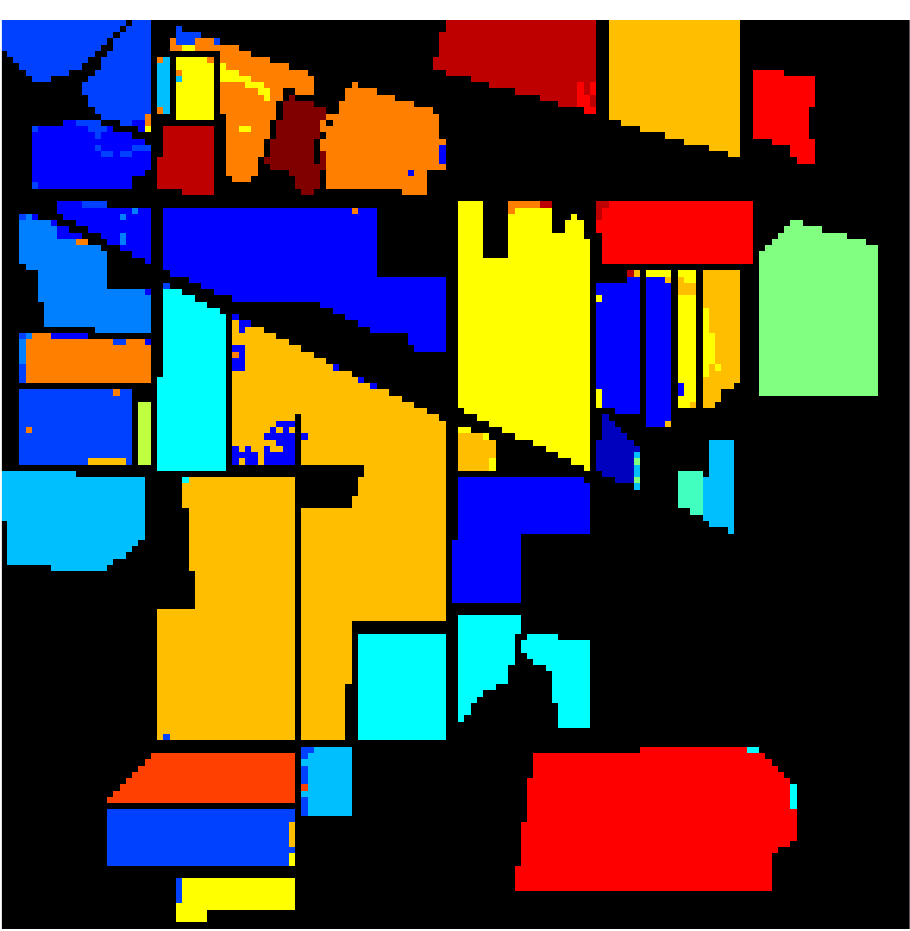}
    \caption{}
    \label{fig:IndianPines_traditional_pc10_label}
  \end{subfigure}
  \hfill
  %
  \begin{subfigure}[b]{\figcbar}
    \centering
    \includegraphics[height=0.985\heightIndianPines]{%
      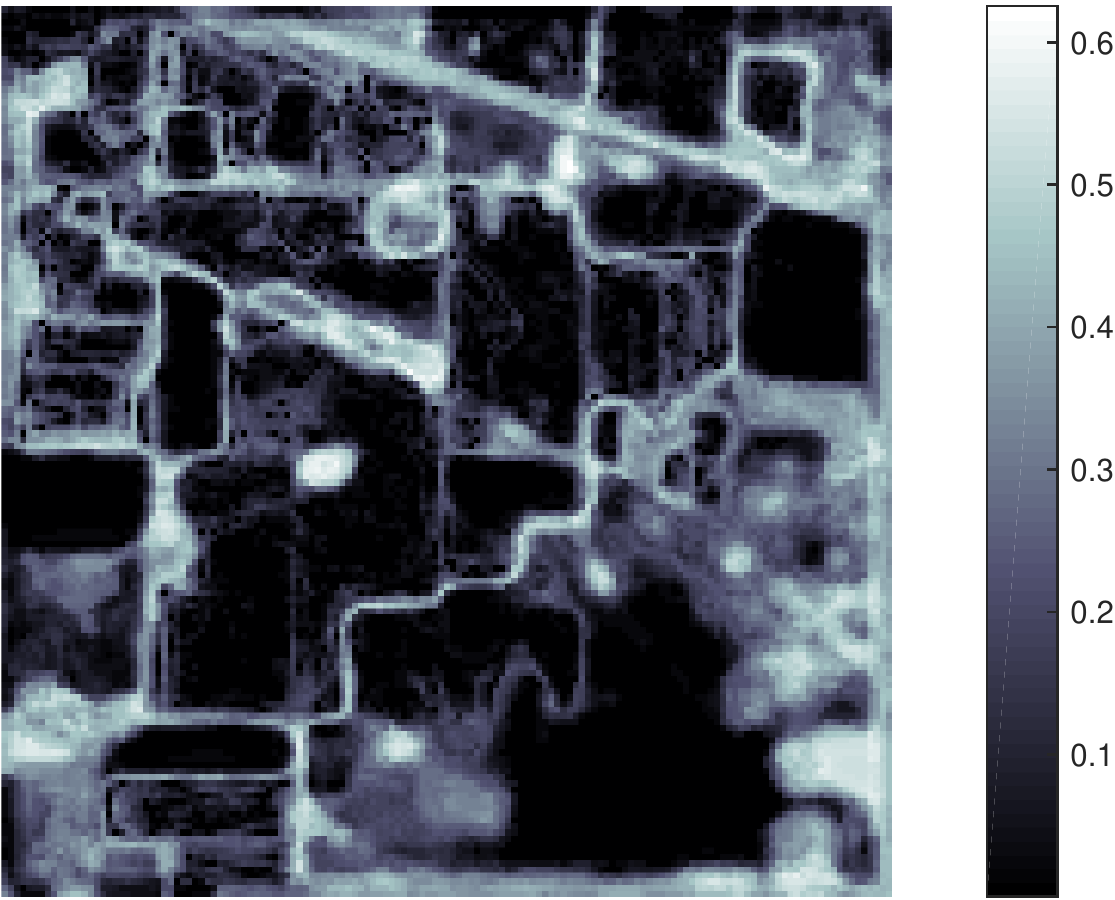}
    \caption{\shiftcbarIndianPines}
    \label{fig:IndianPines_traditional_pc10_clutter}
  \end{subfigure}
  \hfill
  %
  \begin{subfigure}[b]{\fignocbar}
    \centering
    \includegraphics[height=\heightIndianPines]{%
      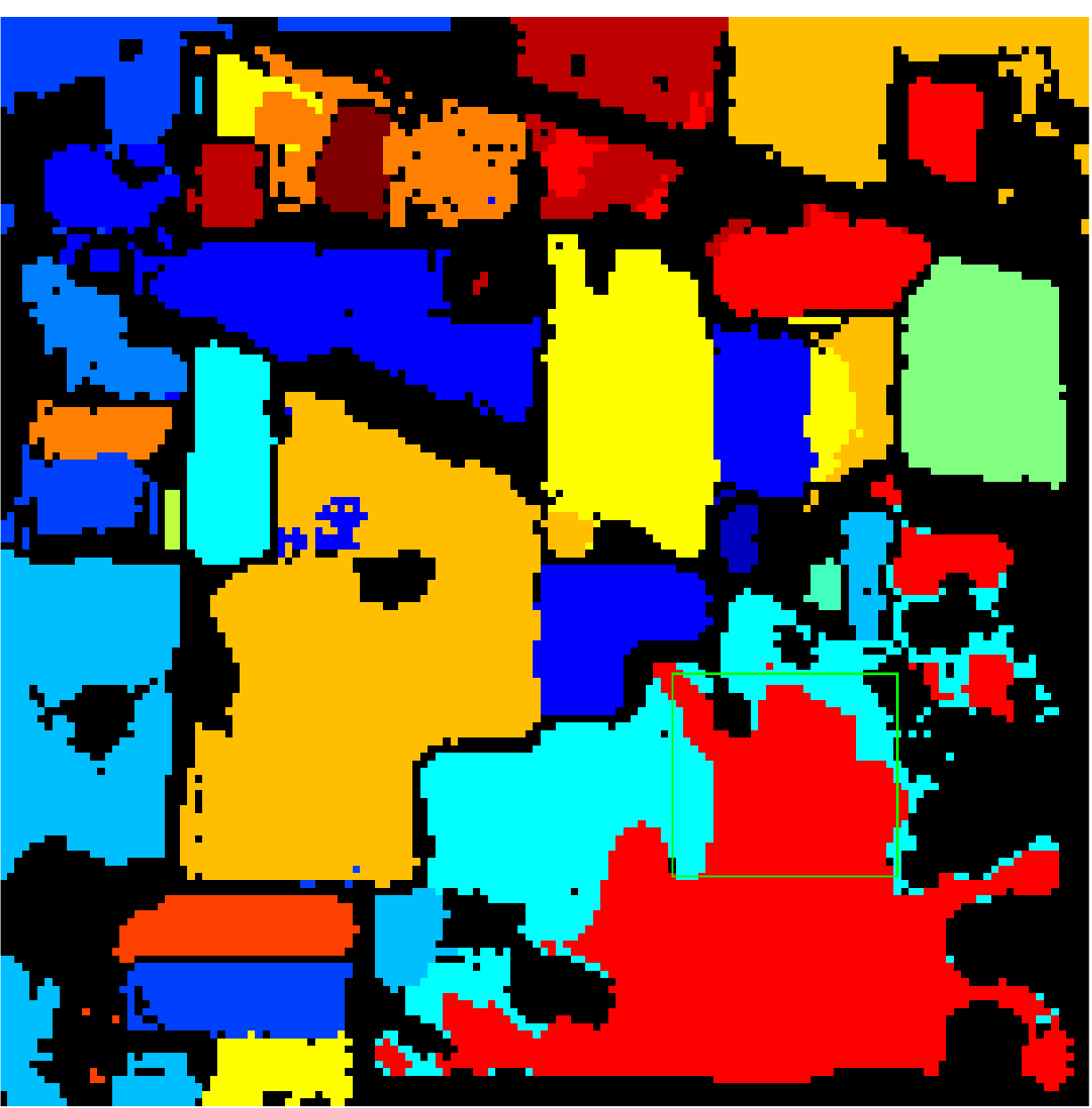}
    \caption{ }
    \label{fig:IndianPines_pc10_label_clutter_masked}
  \end{subfigure}
  \\
  %
  \begin{subfigure}[b]{\fignocbar}
    \centering
    \includegraphics[height=\heightIndianPines]{%
      IndianPines_groundtruth}
    \caption{ }
    \label{fig:IndianPines_traditional_gt}
  \end{subfigure}
  \hfill\hfill
  %
   \begin{subfigure}[b]{\fignocbar}
    \centering
    \includegraphics[height=\heightIndianPines]{%
      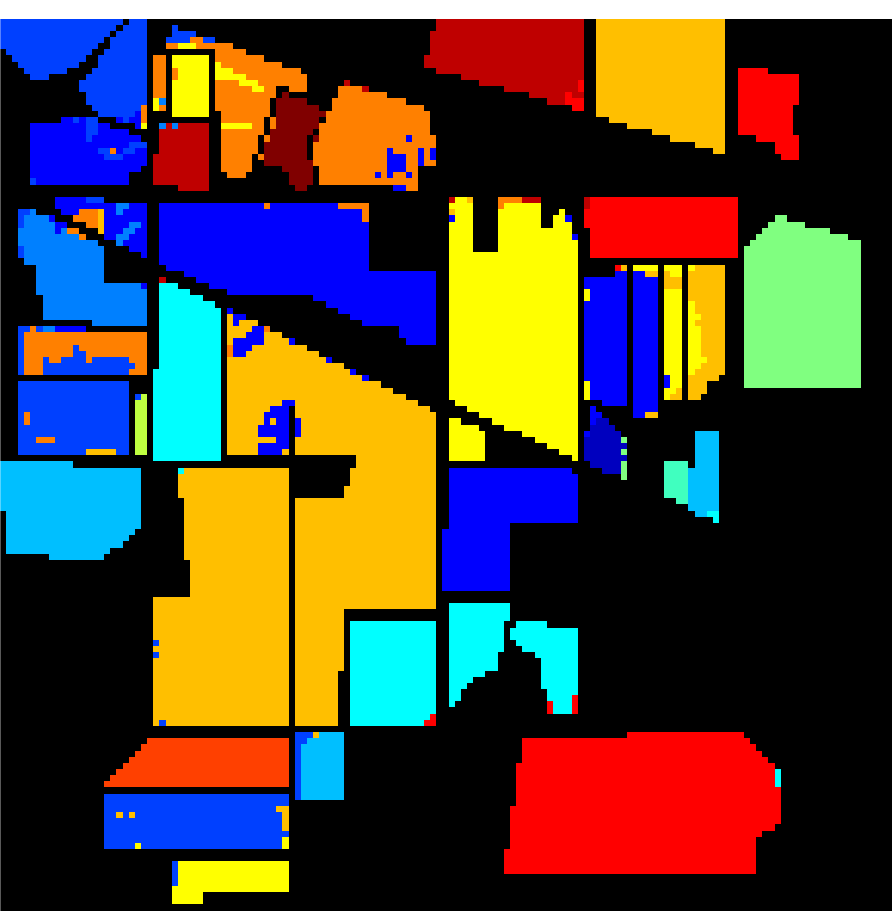}
    \caption{ }
    \label{fig:IndianPines_traditional_pc5_label}
  \end{subfigure}
  \hfill
  %
  \begin{subfigure}[b]{\figcbar}
    \centering
    \includegraphics[height=0.985\heightIndianPines]{%
      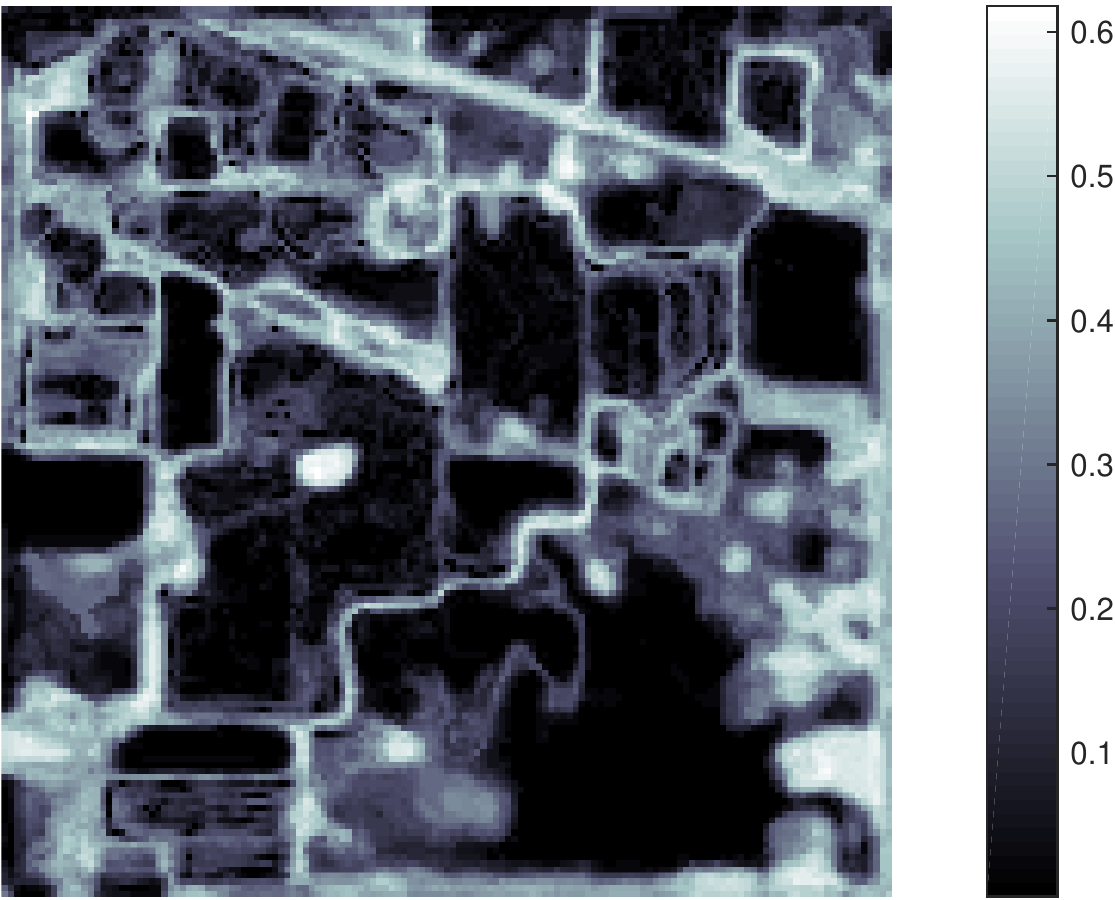}
    \caption{\shiftcbarIndianPines}
    \label{fig:IndianPines_traditional_pc5_clutter}
  \end{subfigure}
  \hfill
  %
  \begin{subfigure}[b]{\fignocbar}
    \centering
    \includegraphics[height=\heightIndianPines]{%
      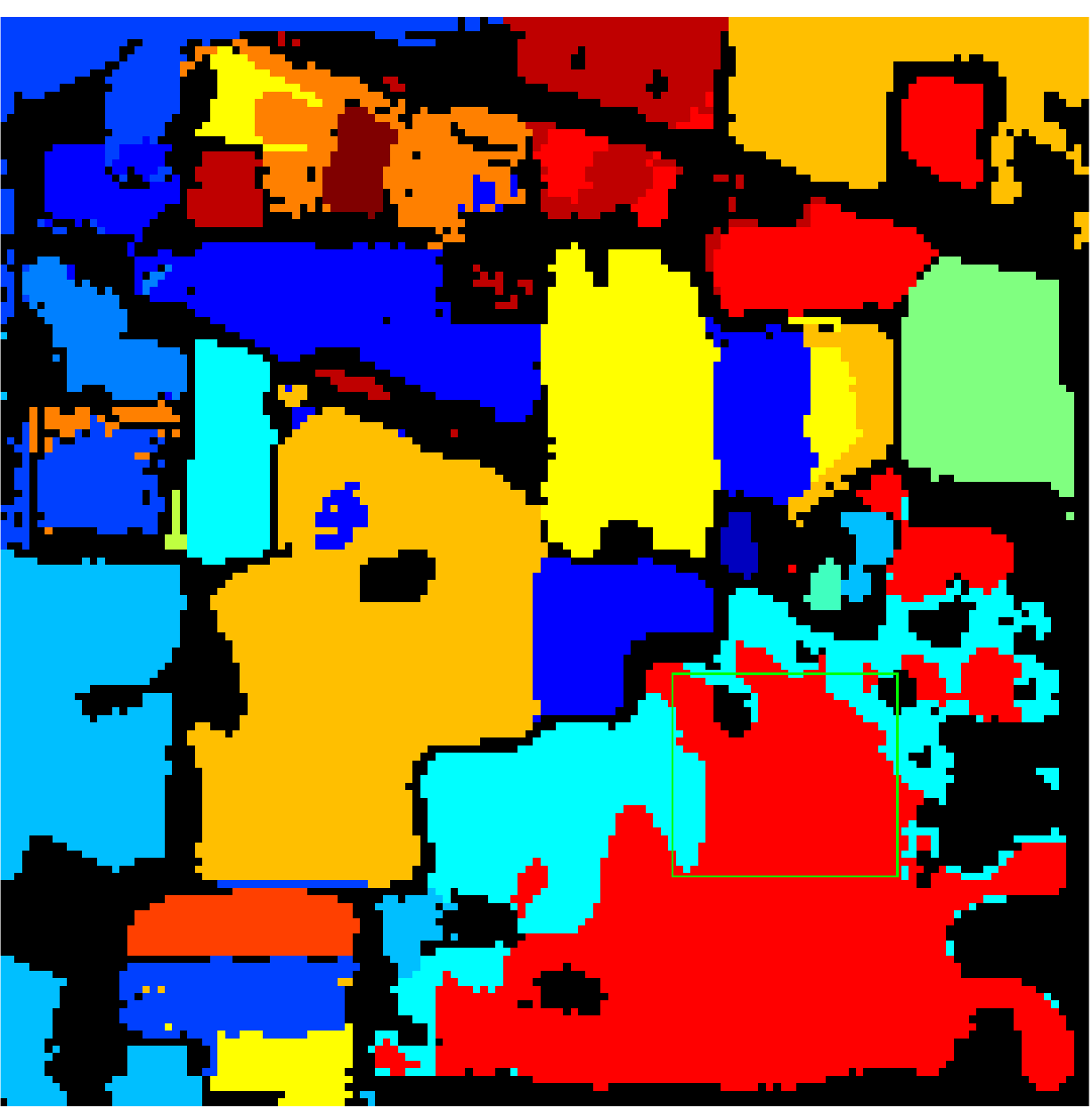}
    \caption{ }
    \label{fig:IndianPines_pc5_label_clutter_masked}
  \end{subfigure}
  \caption{%
    Classification and clutter detection results for the Indian Pines scene.
    \emph{(a) and (e)}~RGB composite~\cite{sotoca_hyperspectral_2006} and
    manual classification labeling and mask.
    \emph{(b)--(d)}~10\% labeled data sampling: masked
    classification; classification entropy map; classification and
    clutter removal with $\thresClutter = 0.25$.
    \emph{(f)--(h)}~5\% labeled data sampling: same as (b)--(d)
    with $\thresClutter = 0.30$.}
  \label{fig:IndianPines_traditional}
\end{figure*}


%
\section{Experiments}
\label{sec:experiments}


\subsection{Datasets}
\label{sec:exp-datasets}


Two publicly available HSI datasets have been used to appraise the
effectiveness of our approach. One is the \emph{Indian Pines}%
\footnote{\url{https://engineering.purdue.edu/~biehl/MultiSpec/hyperspectral.html}}
scene, recorded by the AVIRIS sensor~\cite{green_imaging_1998} in
Northwestern Indiana, USA. It consists mostly of agricultural plots
(alfalfa, corn, oats, soybean, wheat), and forested regions (woods, and
different sub-classes of grass), while a few buildings may also be
seen. Several classes exhibit significant spectral overlap, as they
correspond to the same basic class under different conditions.

The other is the \emph{University of Pavia}%
\footnote{\url{http://www.ehu.eus/ccwintco/index.php?title=Hyperspectral_Remote_Sensing_Scenes}}
scene, recorded by the ROSIS sensor~\cite{holzwarth_hysens_2003} in Pavia,
Italy. It covers an urban environment, with various solid structures
(asphalt, gravel, metal sheets, bitumen, bricks), natural objects (trees,
meadows, soil), and shadows. Objects whose compositions differ from the
labeled ones are considered as clutter.

Both datasets are available with a manually labeled mask, where each pixel
is assigned a class (color) or is discarded as clutter (black). An RGB
composite image and the labeled mask for the two datasets are shown in
\cref{fig:IndianPines_traditional_RGB,fig:IndianPines_traditional_gt,%
  fig:PaviaU_traditional_RGB,fig:PaviaU_traditional_gt}. A summary of
relevant parameters for the two datasets may be found in
\cref{tab:datasets}, and the corresponding reference label maps are shown
in \cref{sec:sup-reference-data}.




%
\begin{figure*}[t]
  \centering
  %
  \begin{subfigure}[b]{\fignocbar}
    \centering
    \includegraphics[height=0.975\heightPaviaU]{%
      PaviaU_RGB_kang_2014}%
    \caption{ }
    \label{fig:PaviaU_traditional_RGB}
  \end{subfigure}
  \hfill\hfill 
  %
  \begin{subfigure}[b]{\fignocbar}
    \centering
    \includegraphics[height=\heightPaviaU]{%
      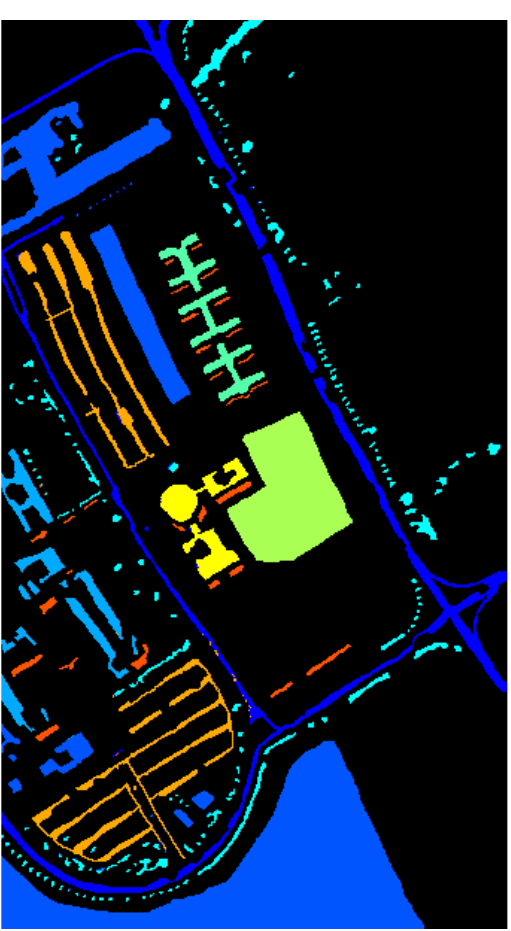}
    \caption{ }
    \label{fig:PaviaU_traditional_pc5_label}
  \end{subfigure}
  \hfill
  %
  \begin{subfigure}[b]{\figcbar}
    \centering
    \includegraphics[height=0.995\heightPaviaU]{%
      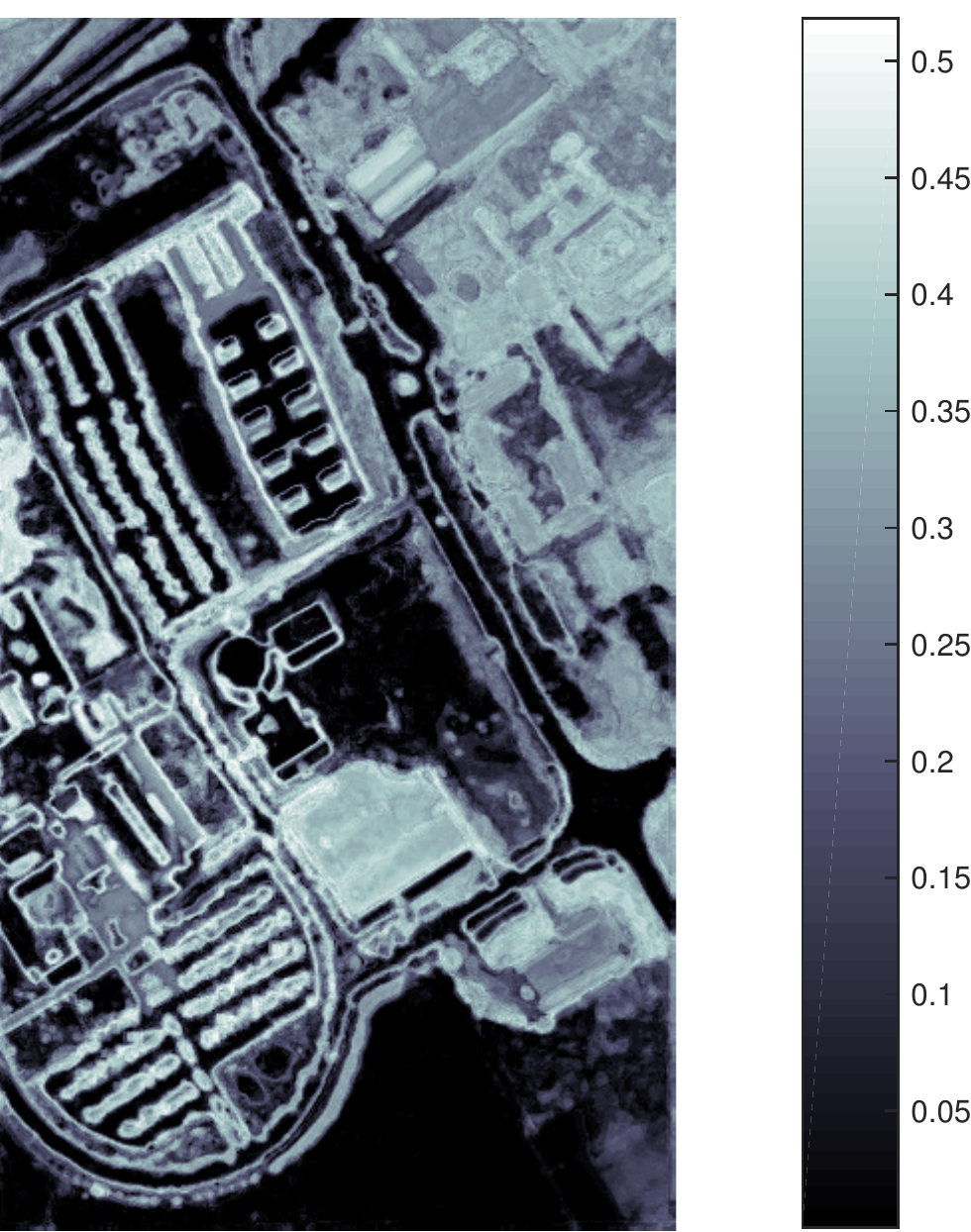}
    \caption{\shiftcbarPaviaU}
    \label{fig:PaviaU_traditional_pc5_clutter}
  \end{subfigure}
  \hfill
  %
  \begin{subfigure}[b]{\fignocbar}
    \centering
    \includegraphics[height=\heightPaviaU]{%
      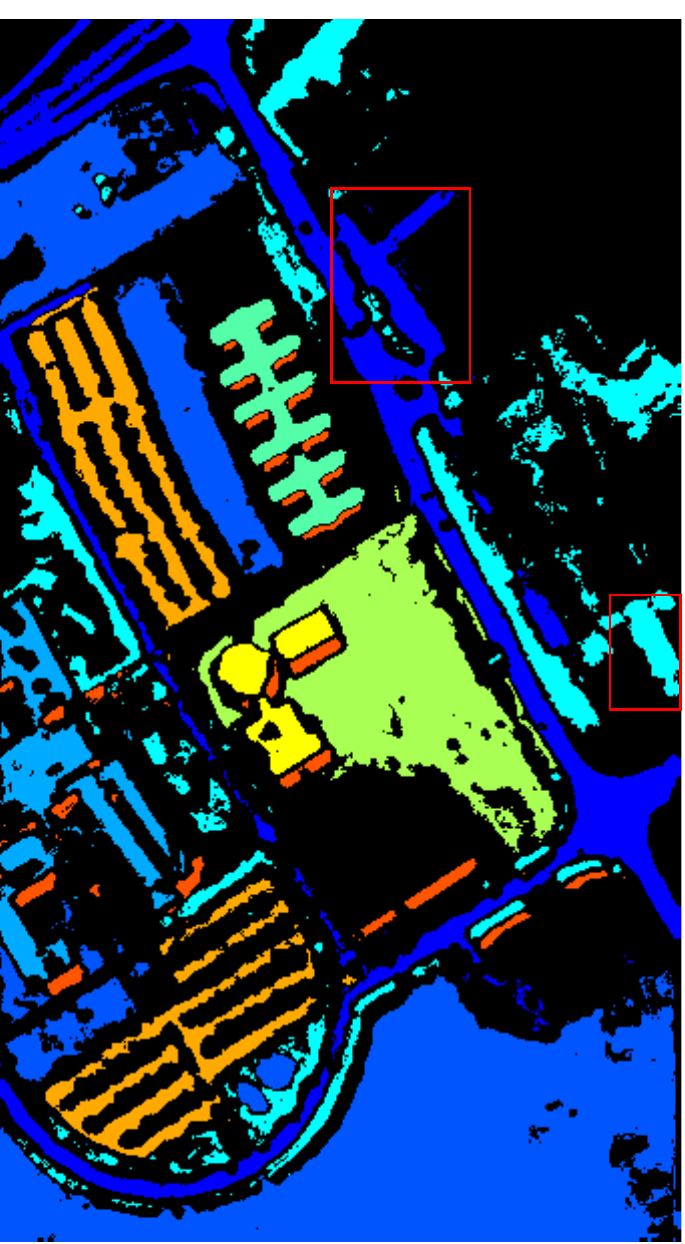}
    \caption{ }
    \label{fig:PaviaU_pc5_label_clutter_masked}
  \end{subfigure}
  \\
  \begin{subfigure}[b]{\fignocbar}
    \centering
    \includegraphics[height=\heightPaviaU]{%
      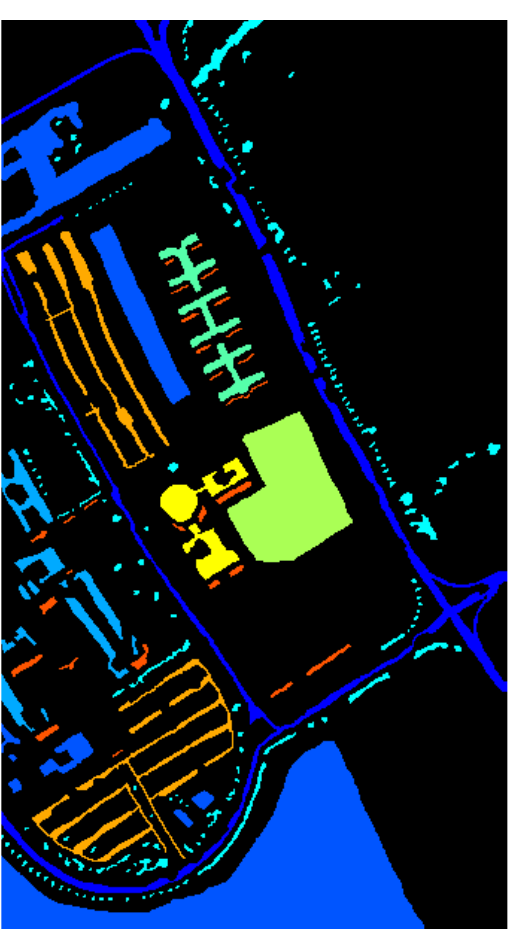}
    \caption{ }
    \label{fig:PaviaU_traditional_gt}
  \end{subfigure}
  \hfill\hfill  
  %
  \begin{subfigure}[b]{\fignocbar}
    \centering
    \includegraphics[height=\heightPaviaU]{%
      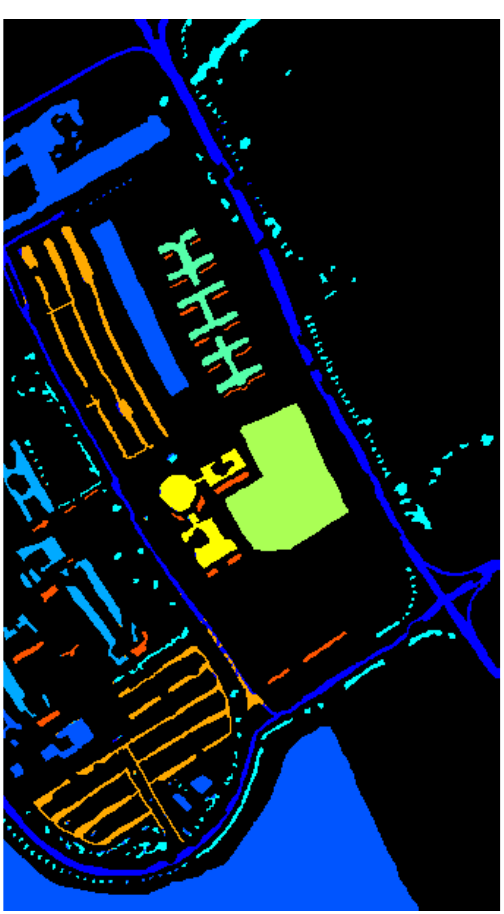}
    \caption{ }
    \label{fig:PaviaU_traditional_pc5_label}
  \end{subfigure}
  \hfill 
  %
  \begin{subfigure}[b]{\figcbar}
    \centering
    \includegraphics[height=0.995\heightPaviaU]{%
      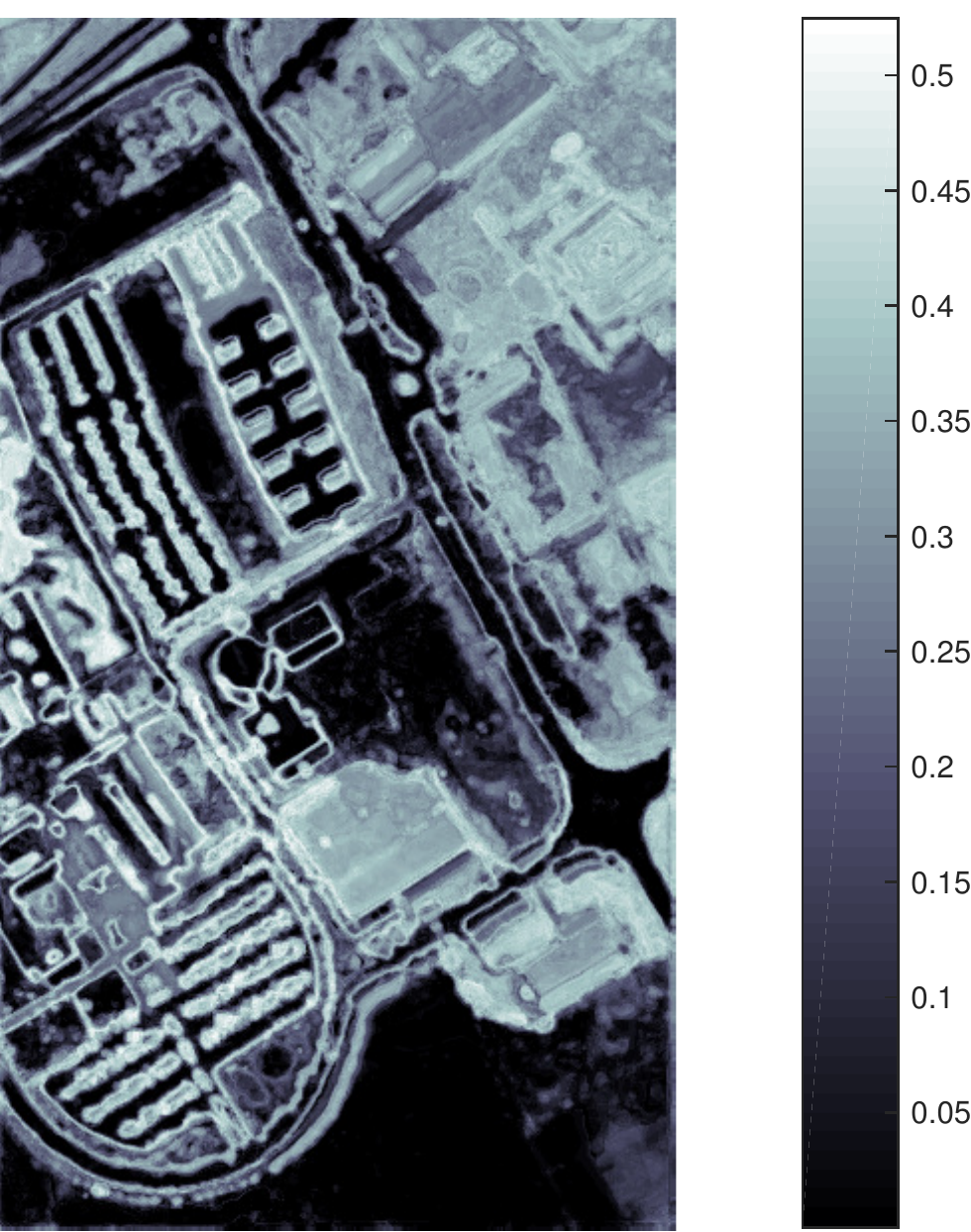}
    \caption{\shiftcbarPaviaU}
    \label{fig:PaviaU_traditional_pc2_clutter}
  \end{subfigure}
  \hfill
  \begin{subfigure}[b]{\fignocbar}
    \centering
    \includegraphics[height=\heightPaviaU]{%
      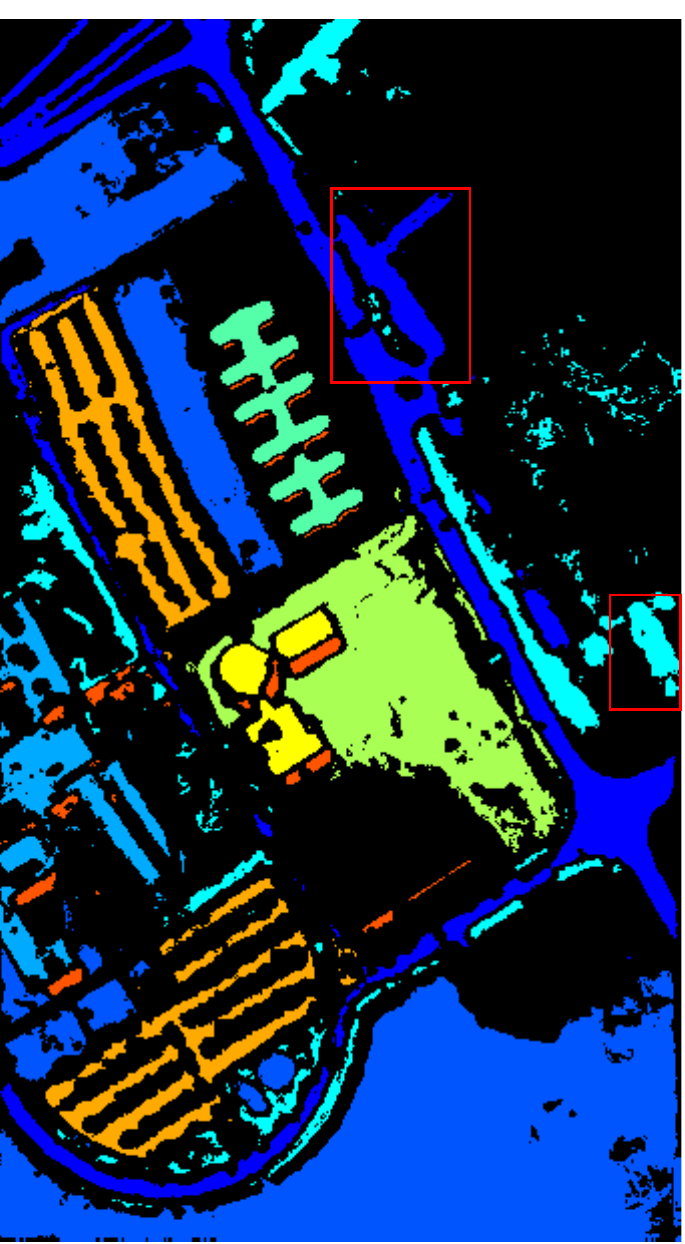}
    \caption{ }
    \label{fig:PaviaU_pc2_label_clutter_masked}
  \end{subfigure}
  \caption{%
    Classification and clutter detection results for the University of
    Pavia scene. 
    \emph{(a) and (e)}~RGB composite~\cite{kang_spectral-spatial_2014} and
    manual classification labeling and mask.
    \emph{(b)--(d)}~5\% labeled data sampling: masked classification;
    classification entropy map; classification and clutter removal with
    $\thresClutter = 0.15$.
    \emph{(f)--(h)}~2\% labeled data sampling: same as (b)--(d) with
    $\thresClutter = 0.15$.}
  \label{fig:PaviaU_traditional}
\end{figure*}



\subsection{Experimental set-up}
\label{sec:exp-setup}


Prior to any other processing, $24$ noisy bands were removed from the
Indian Pines dataset; these correspond to $20$ water absorption
bands~\cite{tadjudin_covariance_1998} and another $4$ that were dominated
by noise. All bands were kept for the University of Pavia dataset (albeit
$12$ have already been removed from the data in the public repository).

For all experiments presented here, the algorithmic ensemble parameters
were as follows:%
\begin{inparaenum}[(i)]
\item The spectral bank consisted of the identity (i.e.\ the original
  attributes were used), numerical gradient, mean, and standard deviation
  filters; spectral features were extracted at two scales using the
  \emph{\{whole, odd, even\}} spectrum.
\item A spatial box filter was applied to all features, using a
  $\patchSize \times \patchSize$ neighborhood, where
  $\patchSize \in \{3,5\}$.
\item The size of local manifold neighborhoods was $\sizeNhood \in \{5, 10,
  15\}$.
\item The dimension of the manifold-representation classification space was
  $\dimMan \in \{10, 20, 30\}$.
\end{inparaenum}

The resulting ensemble was comprised by a single instance for each element
in the Cartesian product of algorithmic parameter sets, for a total of 54
embeddings and low-dimensional representations of the HSI
data. Nearest-neighbor searching was bounded using a $51 \times 51$ sliding
window. Labels were acquired via pixel-wise nearest-neighbor classification
for each instance and non-weighted consensus for the ensemble.

Reference labeled data for classification were sampled uniformly at random,
using the reference labeled mask to extract samples at $10\%$ or $5\%$
density per class for the Indian Pines dataset, and at $5\%$ or $2\%$
density per class for the University of Pavia dataset. The label-set size
ranged from approximately $120$ to just $1$ pixel per class, depending on
the relative coverage of the HSI scene.




\subsection{Rendering schemes}
\label{sec:exp-rendering-metrics}


We render experimental results in three ways. First, we follow the
conventional scheme, where only pixels that belong to a class in the
reference label map are considered---the rest are discarded, regardless of
the corresponding classification results. Quantitative results are provided
using the standard overall accuracy (OA; percentage of correctly classified
pixels) and average accuracy (AA; average of class-wise classification
accuracy percentages) metrics.

While the OA and AA metrics allow comparisons with a reference (manual)
classification result, they cannot capture other aspects of the
classification problem, and provide no information as to the separation of
clutter and labeled samples. Hence, in the absence of available reference
data for the whole scene, we resort to visual appraisal of the
classification and clutter detection results using the other two rendering
schemes.

One is a gray-scale rendering of the classification entropy (clutter
estimate) map; ideally, it should be dark for labeled regions and bright
for clutter.
Last, we render the final classification results with our approach, by
merging the ensemble consensus labeling with a clutter mask, obtained by
thresholding the clutter estimate image. Good results should have the
following qualities: each region is classified correctly, region boundaries
are respected by the classification map, and clutter is accurately
identified.




%
\begin{table*}
  \centering
  \caption{Classification accuracy for the embedding ensemble and
    instances.}
  \label{tab:classification-accuracy}
  \begin{tabular}{%
    c  
    c<{\%}
    r@{.}l @{$\,\pm\,$} r@{.}l @{$\;\,$} >{(}r@{.}l<{)}  
    r@{.}l @{$\,\pm\,$} r@{.}l @{$\;\,$} >{(}r@{.}l<{)}  
    >{\hshiftacc}r@{.}l  
    >{\hshiftacc}r@{.}l
    }
    \toprule
      \multirow{2}{*}[\vshiftmrow]{\textbf{Dataset}}
    & \multicolumn{1}{c}{%
      \multirow{2}{*}[\vshiftmrow]{\parbox{5em}{%
      \centering \bfseries Labeled samples}}}
    & \multicolumn{12}{c}{\textbf{Instances} [mean $\pm$ std (max)]}
    & \multicolumn{4}{c}{\textbf{Ensemble}}
    \\
    \cmidrule(lr){3-14}
    \cmidrule(lr){15-18}
    & \multicolumn{1}{c}{}
    & \multicolumn{6}{c}{\textbf{OA}\ (\%)}
    & \multicolumn{6}{c}{\textbf{AA}\ (\%)}
    & \multicolumn{2}{c}{\textbf{OA}\ (\%)}
    & \multicolumn{2}{c}{\textbf{AA}\ (\%)}
    \\
    \midrule
    \multirow{2}{*}{\parbox{4em}{\centering Indian Pines}}
    & 5
    & 85&79   &   4&12   &   92&88
    & 82&06   &   5&68   &   91&66
    & 95&39
    & 94&85
    \\
    & 10
    & 90&00   &   3&57   &   96&07
    & 87&68   &   4&87   &   95&45
    & 97&34
    & 97&13
    \\
    \midrule[0.1pt]
    \multirow{2}{*}{\parbox{4em}{\centering University of Pavia}}
    & 2
    & 94&87   &   2&38   &   97&86
    & 92&29   &   3&68   &   97&13
    & 98&84
    & 98&42
    \\
    & 5
    & 96&92   &   1&76   &   98&99
    & 95&41   &   2&47   &   98&43
    & 99&60
    & 99&32
    \\
    \bottomrule
  \end{tabular}
\end{table*}



\subsection{Results}
\label{sec:exp-results}


A summary of the classification accuracy metrics for both HSI datasets,
measured with respect to the corresponding manually labeled mask, is shown
in \cref{tab:classification-accuracy}, for the embedding
instances as well as the ensemble. We can see that the ensemble outperforms
all instances, having a significant margin from the majority of the
latter. This is especially true for the Indian Pines dataset, which proves
to be more difficult than the University of Pavia one, due to the spectral
overlap between different classes and very low spatial resolution, which
means that there may be substantial variability among pixels of the same
class.
For both datasets, very high classification accuracy is attained. Note,
however, that these metrics only take a portion of the image into account.

%
Results for the Indian Pines dataset are displayed in
\cref{fig:IndianPines_traditional}. It can be readily seen in
\cref{fig:IndianPines_traditional_pc10_label,%
  fig:IndianPines_traditional_pc5_label} that classification errors are mostly
localized around a couple of difficult regions. Nevertheless, the clutter
estimate maps clearly capture the outline structure of the scene---and many
of the mis-classified regions are acknowledged as somewhat
ambiguous. Looking at the fused classification-clutter images in
\cref{fig:IndianPines_pc10_label_clutter_masked,%
  fig:IndianPines_traditional_pc5_clutter}, we can already see the efficacy
of the proposed methodology: the overall structure of the manual label-mask
is recovered nicely, albeit without particularly sharp features. In
addition, we are able to recover regions that were not labeled, although
they rather clearly extend beyond the manually drawn boundaries: for
example, notice the woods area (red) towards the bottom-right corner,
highlighted with a superimposed rectangle.

%
Corresponding results for the University of Pavia dataset are shown in
\cref{fig:PaviaU_traditional}. Here, we attain near-perfect classification
results when compared to the manual labeling. More importantly, however, we
seem to be able to recover a very high-fidelity profile of the whole scene,
without any prior assumptions about the distribution of clutter
pixels. Indeed, objects belonging to labeled classes are identified inside
unlabeled regions, and \cref{fig:PaviaU_traditional_pc5_clutter,%
  fig:PaviaU_pc2_label_clutter_masked} appear to provide a much more
accurate view of the scene than even the manually labeled mask. For
example, two such regions are highlighted, where a stretch of road and a
set of trees are identified in the unlabeled regions, reflecting the view
of the composite color image with high fidelity.
While it can be seen that \cref{fig:PaviaU_pc5_label_clutter_masked} does
perform better than \cref{fig:PaviaU_traditional_pc2_clutter}, it is
noteworthy that the vast majority of the scene structure is recovered using
reference samples with $2\%$ density.



%
\section{Discussion}
\label{sec:discussion}


We have presented a new approach for HSI classification and clutter
detection via employing an algorithmic ensemble of structural feature
embeddings, nonlinear dimension reduction with the LLE method, and a
classifier to be used in the low-dimensional manifold parameter space.
For feature embedding, we have used only a few simple types of feature
transform functions to explore and exploit the spatial and spectral
coherence structure in the HSI data.
These simple steps, following the isometric principles of manifold
structures, have rendered remarkable results for the two datasets studied
in this paper, while each step may be easily modified or customized to suit
a particular application context, if necessary.
Presently, the parameters ranges for manifold dimension estimation and the
number of neighbors are prescribed. A desirable extension is to have such
ranges determined automatically and adaptively for each dataset.

We have given our rationale for utilizing LLE at the core of our approach.
The LLE method can be connected to multiple methods for classification,
segmentation, or clustering. While various extensions to LLE and
alternative, related approaches to manifold derivation exist, we have found
LLE to be as good as or superior to them, while offering a particularly
simple computational structure. There is still more to be understood
regarding behavior of these methods and their connections to one another.




\appendix

%
\section{Reference labeling for the HSI datasets}
\label{sec:sup-reference-data}


The reference classification data (typically used as ground truth) for
the Indian Pines%
\footnote{\url{https://engineering.purdue.edu/~biehl/MultiSpec/hyperspectral.html}}
and University of Pavia%
\footnote{\url{http://www.ehu.eus/ccwintco/index.php?title=Hyperspectral_Remote_Sensing_Scenes}}
scenes are shown in \cref{fig:reference-data-indianpines,%
  fig:reference-data-pavia}.
The unlabeled regions account for $51\%$ of the entire image domain for the
former, and $80\%$ for the latter---see \cref{tab:datasets}.

%
\begin{figure*}[h]
  \centering
  \begin{subfigure}{0.36\linewidth}
    \centering
    \includegraphics[height=\heightIndianPinesBig]{IndianPines_groundtruth}
    \caption{ }
    \label{fig:reference-data-indianpines-mask}
  \end{subfigure}
  \hfill
  \begin{subfigure}{0.60\linewidth}
    \centering
    \includegraphics[height=0.95\heightIndianPinesBig]{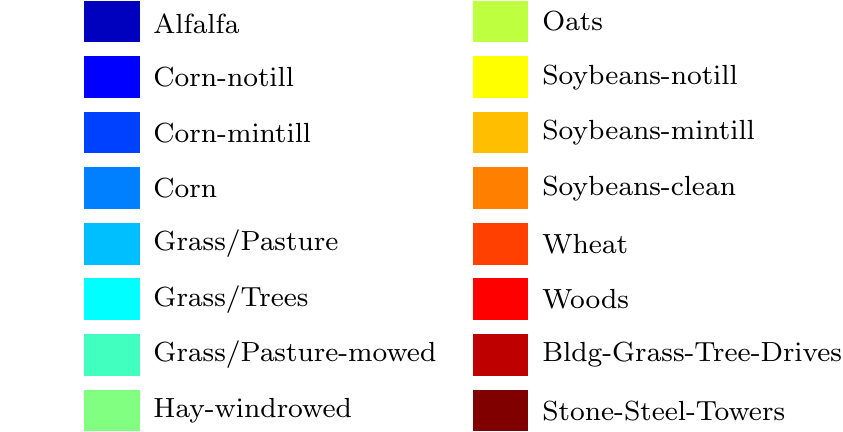}
    \caption{ }
    \label{fig:reference-data-indianpines-legend}
  \end{subfigure}
  \caption{%
    Available labeling information for the Indian Pines scene.
    \emph{(a)} Reference labeling map ($16$ colored classes and black
    unlabeled regions).
    \emph{(b)} Class color legend.%
  }
  \label{fig:reference-data-indianpines}
\end{figure*}

%
\begin{figure*}[h]
  \centering
  \begin{subfigure}{0.36\linewidth}
    \centering
    \includegraphics[height=\heightPaviaUBig]{PaviaU_groundtruth}
    \caption{ }
    \label{fig:reference-data-pavia-mask}
  \end{subfigure}
  \hfill
  \begin{subfigure}{0.60\linewidth}
    \centering
    \includegraphics[height=0.95\heightPaviaUBig]{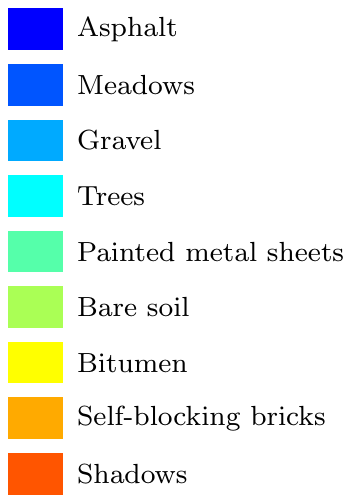}
    \caption{ }
    \label{fig:reference-data-pavia-legend}
  \end{subfigure}
  \caption{%
    Available labeling information for the University of Pavia scene.
    \emph{(a)} Reference labeling map ($9$ colored classes and black
    unlabeled regions).
    \emph{(b)} Class color legend.%
  }
  \label{fig:reference-data-pavia}
\end{figure*}


%
\section{Experiments without feature embedding}
\label{sec:sup-results-original-attributes}


We have presented a comparison of classification results between the
embedding instances and ensemble in
\cref{tab:classification-accuracy-attributes}. In addition to the superior
classification accuracy, the ensemble scheme also enables the provision of
the clutter map.  Here, we provide experimental results that factor out and
highlight the effect of feature space embedding prior to employment of the
LLE method.

\begin{figure*}[p]
\centering
  %
  \begin{subfigure}[b]{\fignocbar}
    \centering
    \includegraphics[height=0.985\heightIndianPines]{%
      IndianPines_RGB_lsi_uji_es}%
    \caption{ }
    \label{fig:IndianPines_noembedding_RGB}
  \end{subfigure}
  \hfill\hfill
  %
  \begin{subfigure}[b]{\fignocbar}
    \centering
    \includegraphics[height=\heightIndianPines]{%
      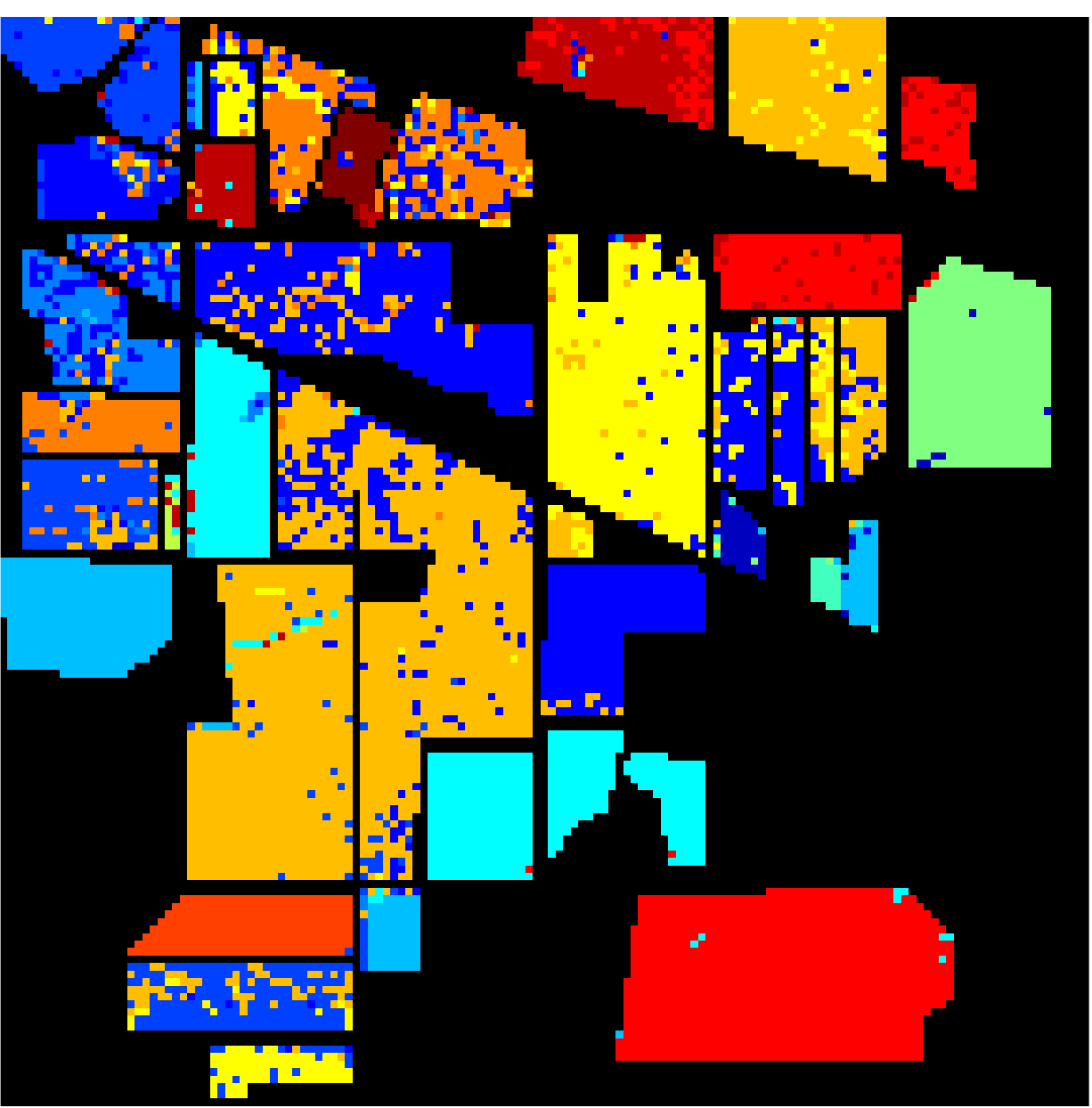}
    \caption{}
    \label{fig:IndianPines_noembedding_pc10_label}
  \end{subfigure}
  \hfill
  %
  \begin{subfigure}[b]{\figcbar}
    \centering
    \includegraphics[height=0.985\heightIndianPines]{%
      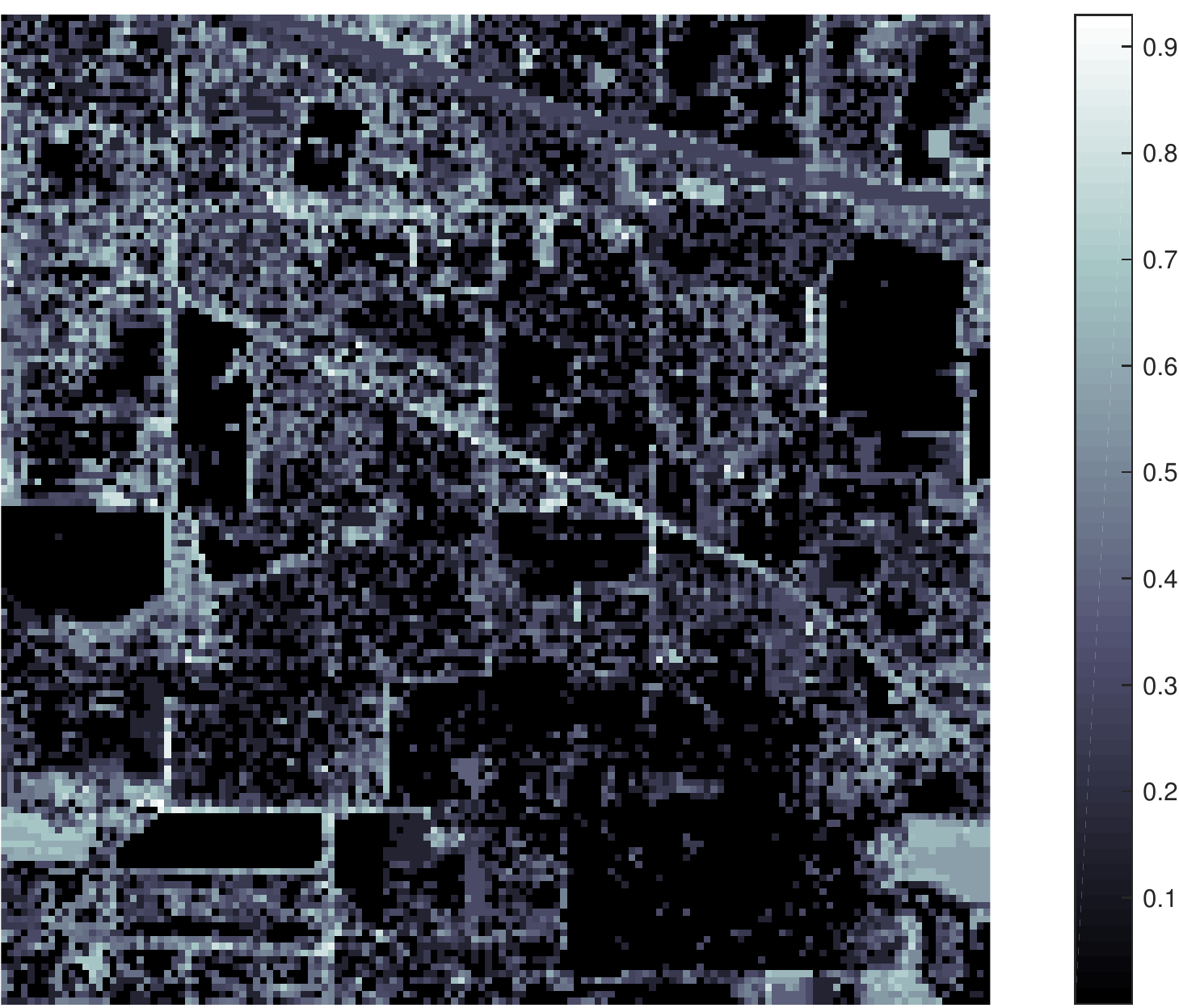}
    \caption{\shiftcbarIndianPines}
    \label{fig:IndianPines_noembedding_pc10_clutter}
  \end{subfigure}
  \hfill
  %
  \begin{subfigure}[b]{\fignocbar}
    \centering
    \includegraphics[height=\heightIndianPines]{%
      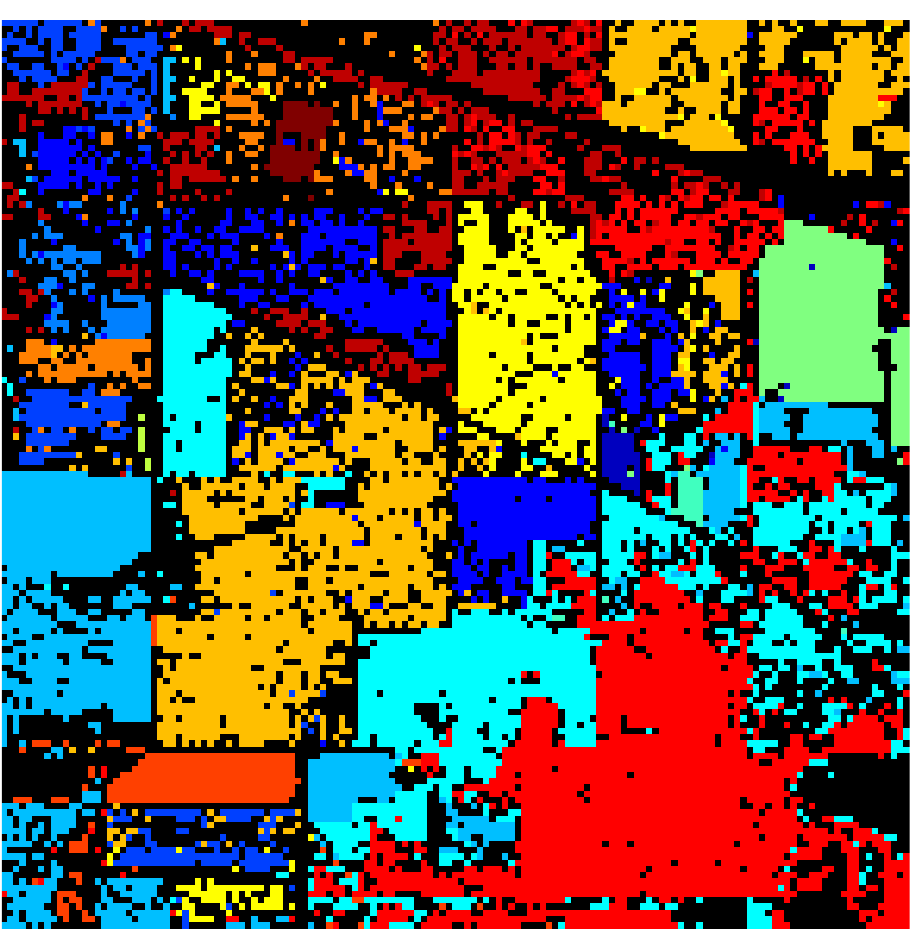}
    \caption{ }
    \label{fig:IndianPines_noembedding_pc10_label_clutter_masked}
  \end{subfigure}
  \\
  %
  \begin{subfigure}[b]{\fignocbar}
    \centering
    \includegraphics[height=\heightIndianPines]{%
      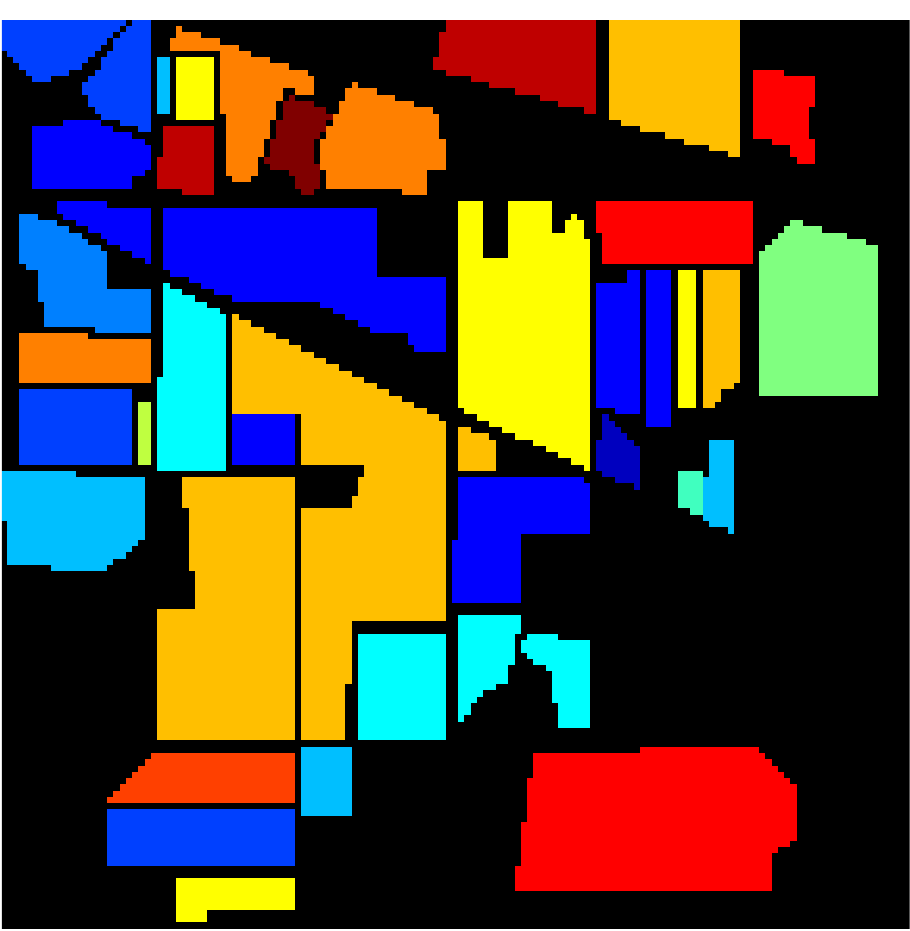}
    \caption{ }
    \label{fig:IndianPines_noembedding_gt}
  \end{subfigure}
  \hfill\hfill
  %
   \begin{subfigure}[b]{\fignocbar}
    \centering
    \includegraphics[height=\heightIndianPines]{%
      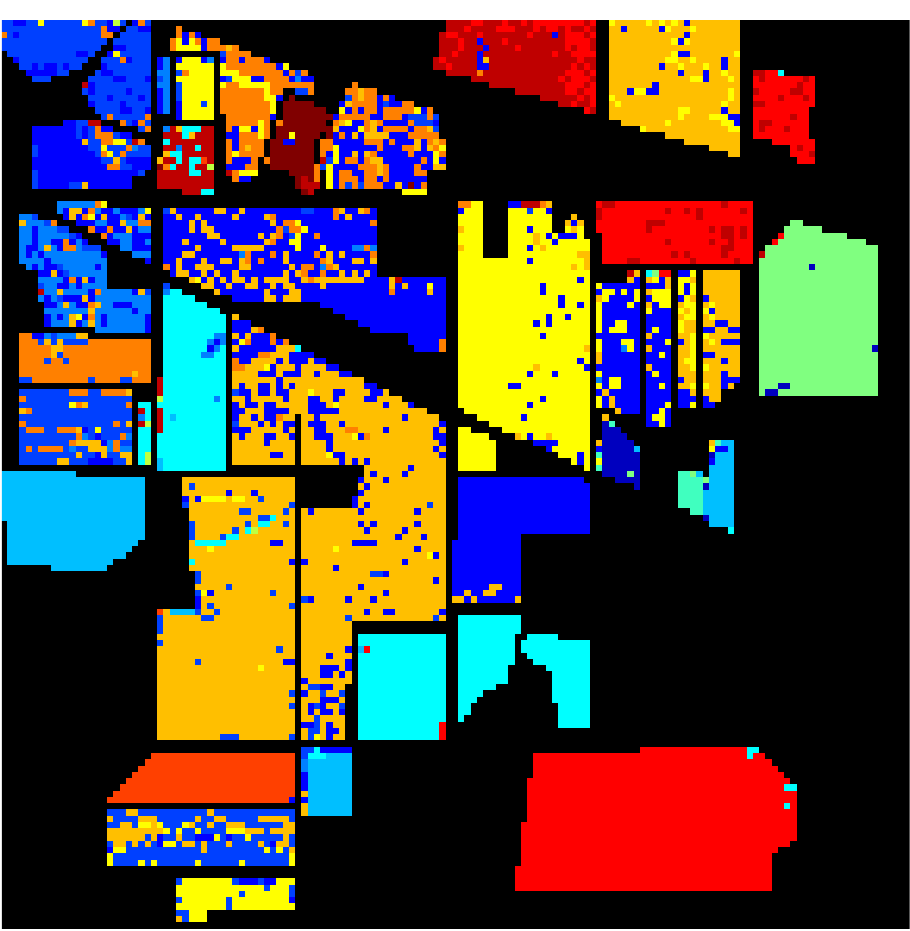}
    \caption{ }
    \label{fig:IndianPines_noembedding_pc5_label}
  \end{subfigure}
  \hfill
  %
  \begin{subfigure}[b]{\figcbar}
    \centering
    \includegraphics[height=0.985\heightIndianPines]{%
      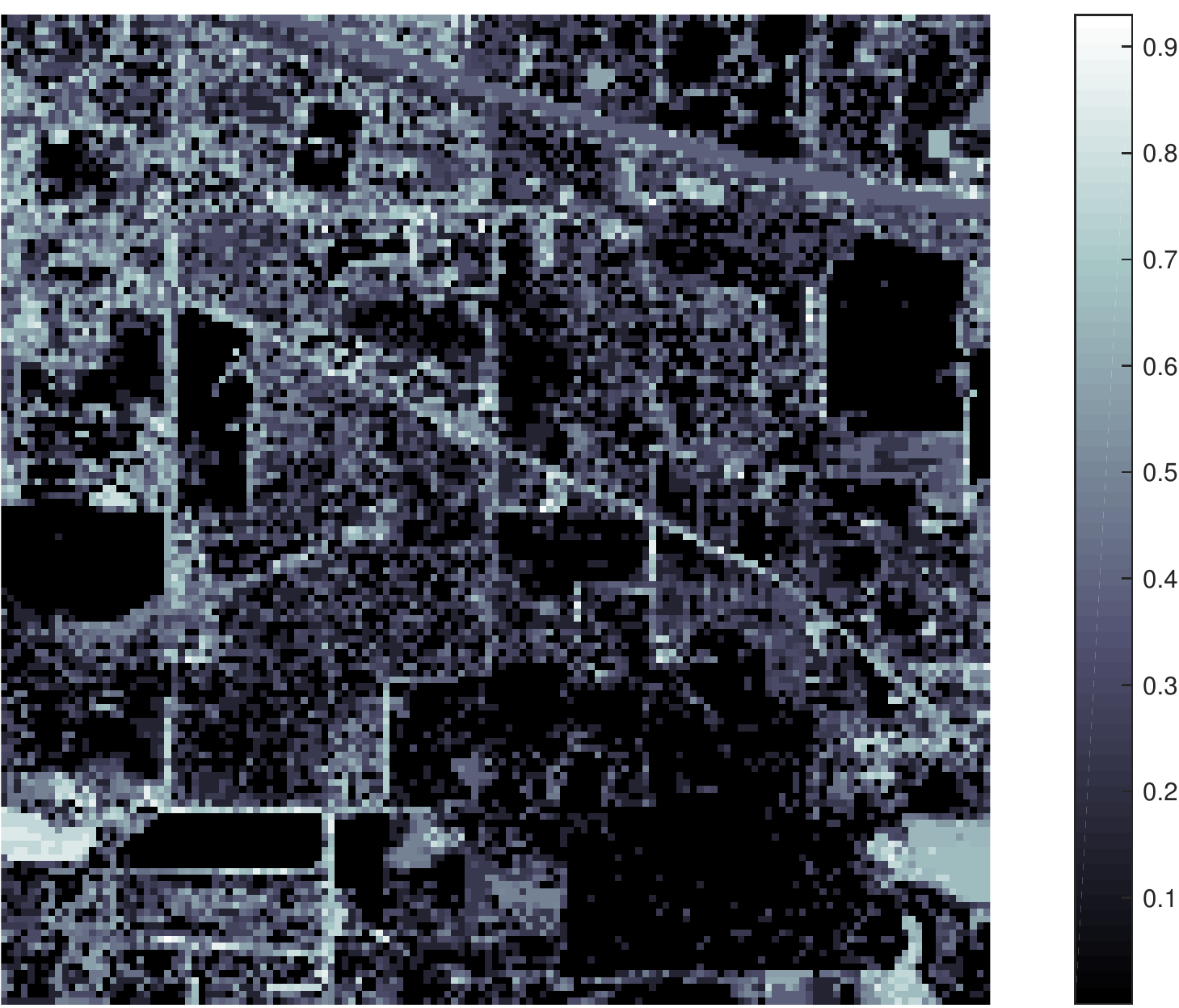}
    \caption{\shiftcbarIndianPines}
    \label{fig:IndianPines_noembedding_pc5_clutter}
  \end{subfigure}
  \hfill
  %
  \begin{subfigure}[b]{\fignocbar}
    \centering
    \includegraphics[height=\heightIndianPines]{%
      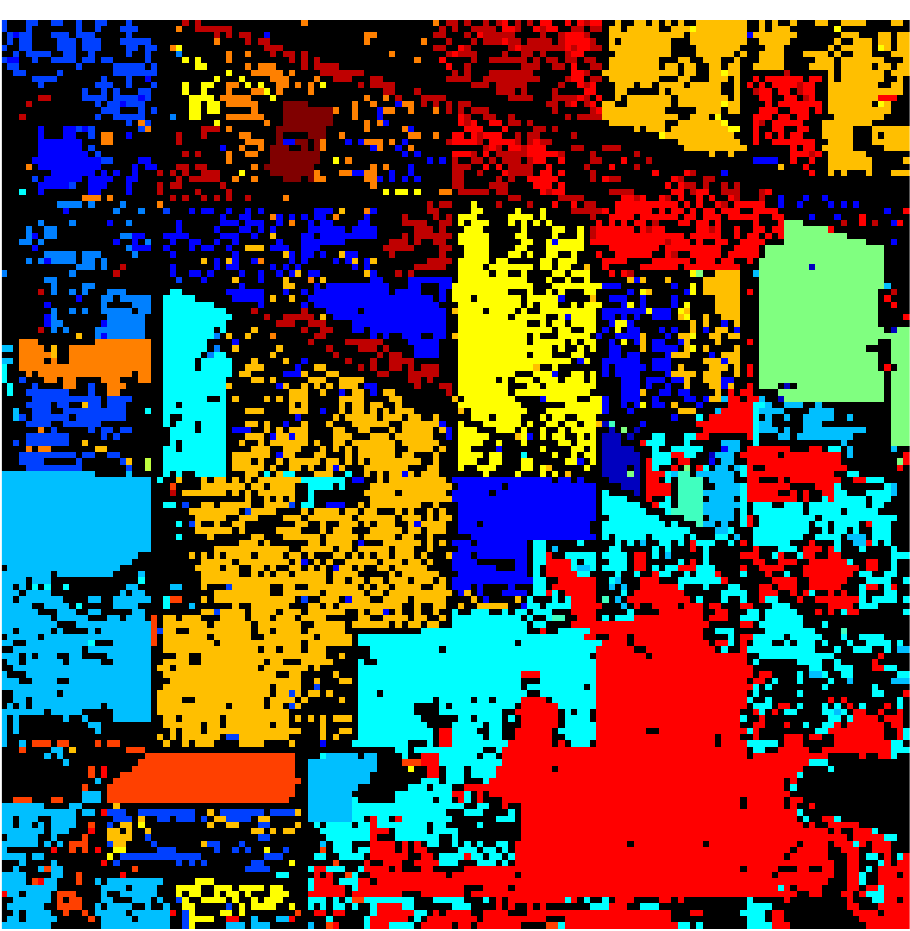}
    \caption{ }
    \label{fig:IndianPines_noembedding_pc5_label_clutter_masked}
  \end{subfigure}
  \caption{%
    Classification and clutter detection results for the Indian Pines scene
    without feature embedding.
    \emph{(a) and (e)}~RGB composite~\cite{sotoca_hyperspectral_2006} and
    manual classification labeling and mask.
    \emph{(b)--(d)}~10\% labeled data sampling: masked
    classification; classification entropy map; classification and
    clutter removal with $\thresClutter = 0.25$.
    \emph{(f)--(h)}~5\% labeled data sampling: same as (b)--(d) with
    $\thresClutter = 0.30$.}
  \label{fig:results-indianpines-attributes}
\end{figure*}

\begin{figure*}[p]
  \centering
%
  \begin{subfigure}[b]{\fignocbar}
    \centering
    \includegraphics[height=0.975\heightPaviaU]{%
      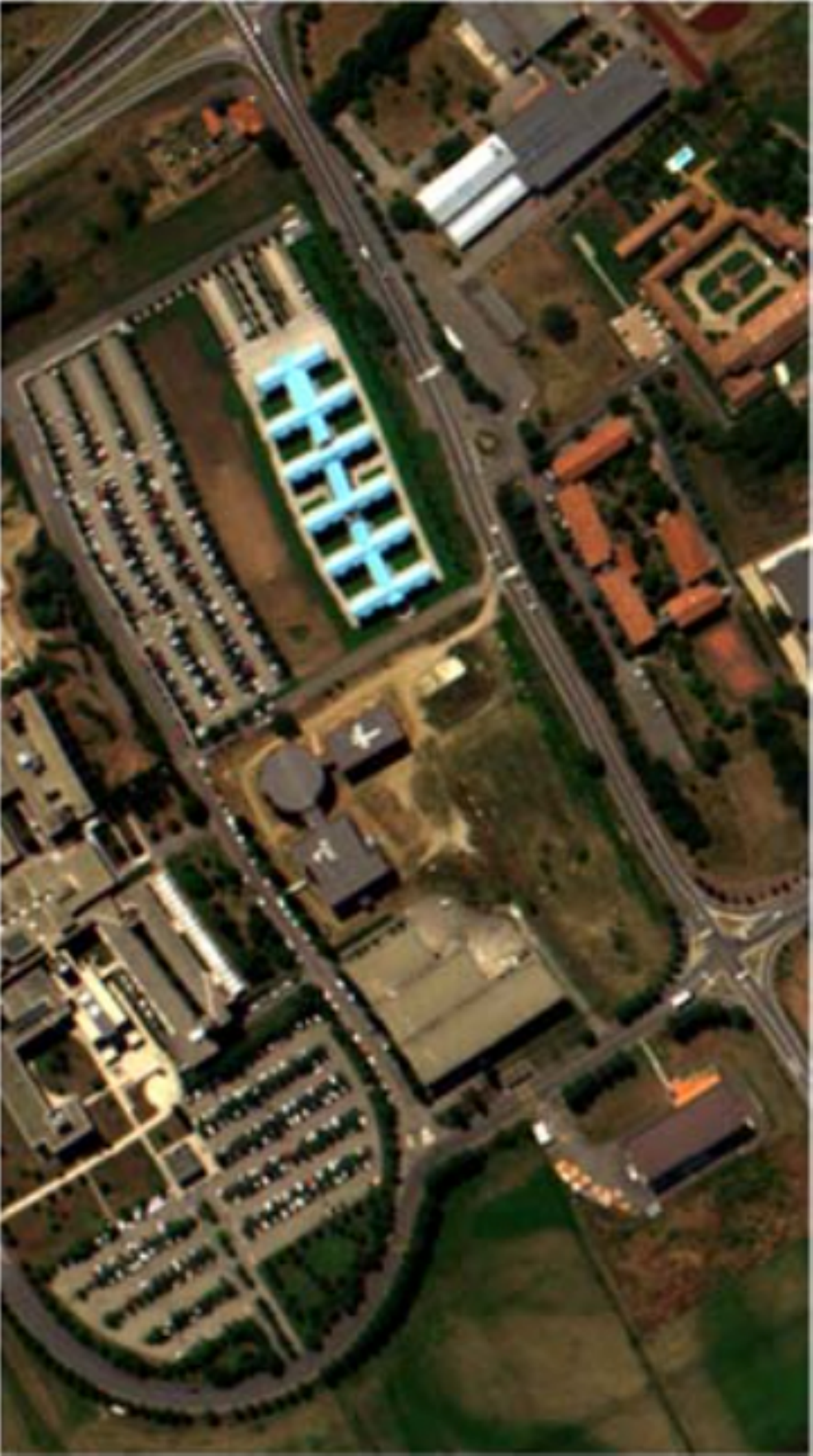}%
    \caption{ }
    \label{fig:PaviaU_noembedding_RGB}
  \end{subfigure}
  \hfill\hfill 
  %
  \begin{subfigure}[b]{\fignocbar}
    \centering
    \includegraphics[height=\heightPaviaU]{%
      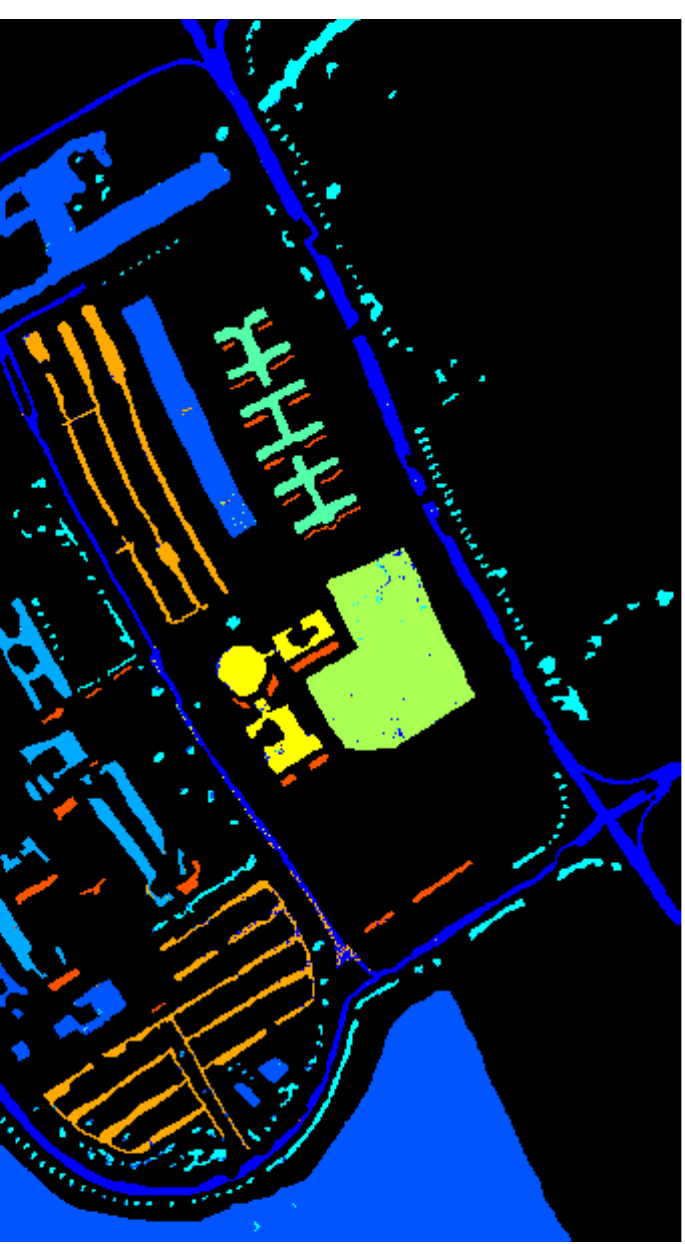}
    \caption{ }
    \label{fig:PaviaU_noembedding_pc5_label}
  \end{subfigure}
  \hfill
  %
  \begin{subfigure}[b]{\figcbar}
    \centering
    \includegraphics[height=0.995\heightPaviaU]{%
      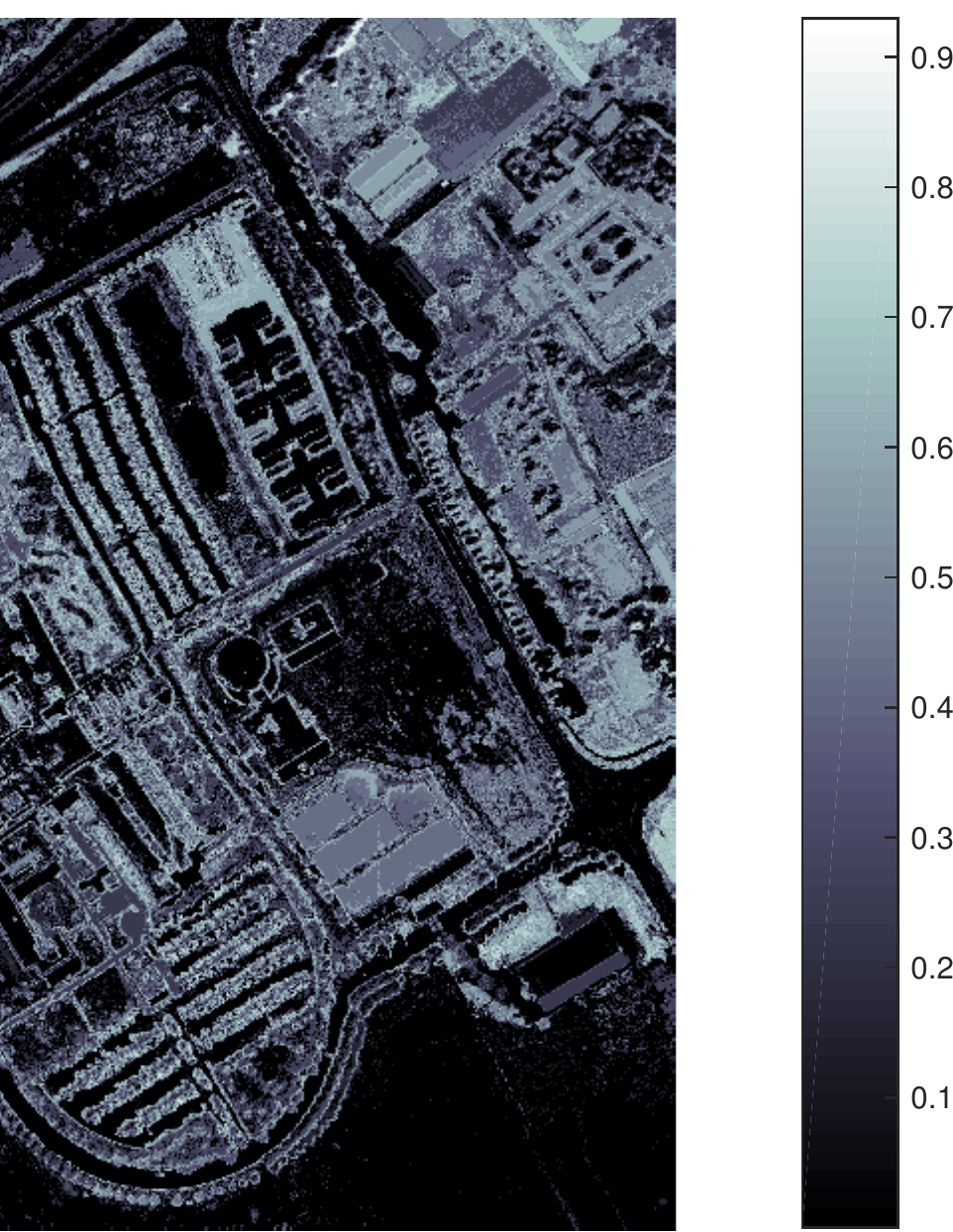}
    \caption{\shiftcbarPaviaU}
    \label{fig:PaviaU_noembedding_pc5_clutter}
  \end{subfigure}
  \hfill
  %
  \begin{subfigure}[b]{\fignocbar}
    \centering
    \includegraphics[height=\heightPaviaU]{%
      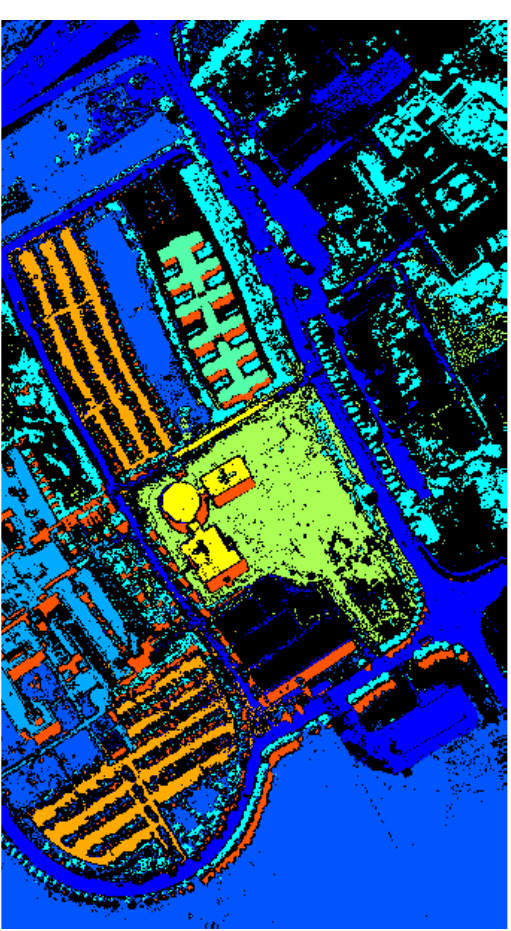}
    \caption{ }
    \label{fig:PaviaU_noembedding_pc5_label_clutter_masked}
  \end{subfigure}
  \\
  \begin{subfigure}[b]{\fignocbar}
    \centering
    \includegraphics[height=\heightPaviaU]{%
      PaviaU_groundtruth.pdf}
    \caption{ }
    \label{fig:PaviaU_noembedding_gt}
  \end{subfigure}
  \hfill\hfill  
  %
  \begin{subfigure}[b]{\fignocbar}
    \centering
    \includegraphics[height=\heightPaviaU]{%
      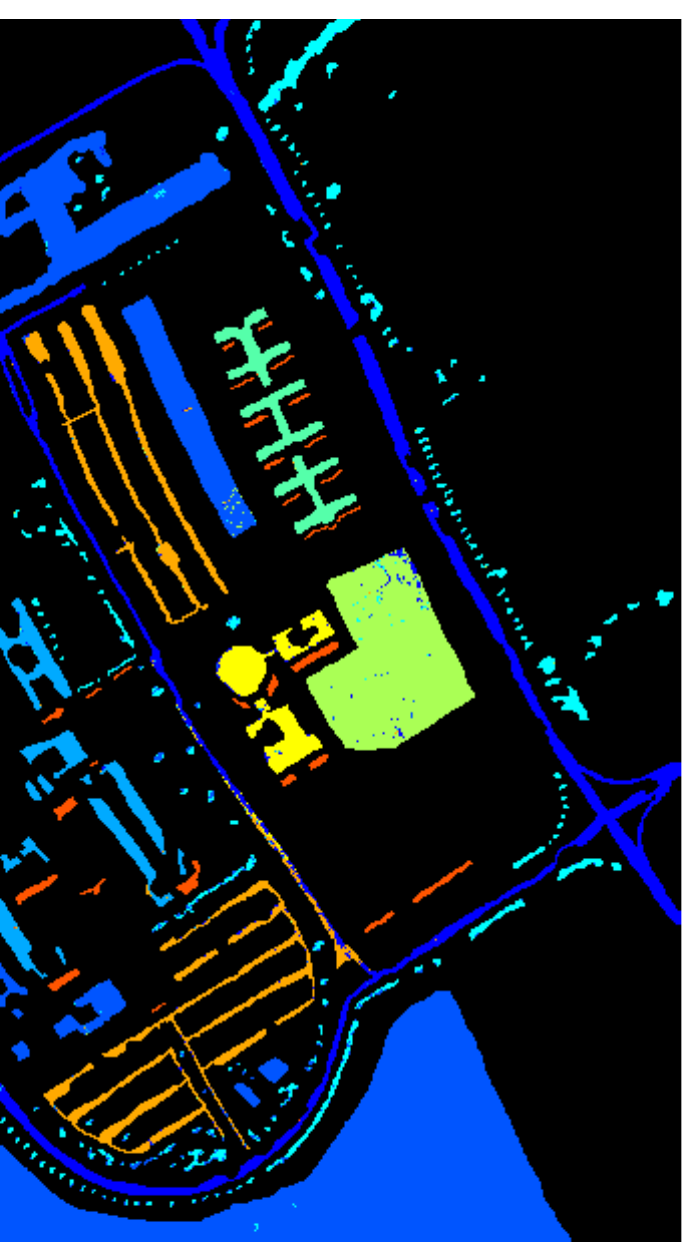}
    \caption{ }
    \label{fig:PaviaU_noembedding_pc5_label}
  \end{subfigure}
  \hfill 
  %
  \begin{subfigure}[b]{\figcbar}
    \centering
    \includegraphics[height=0.995\heightPaviaU]{%
      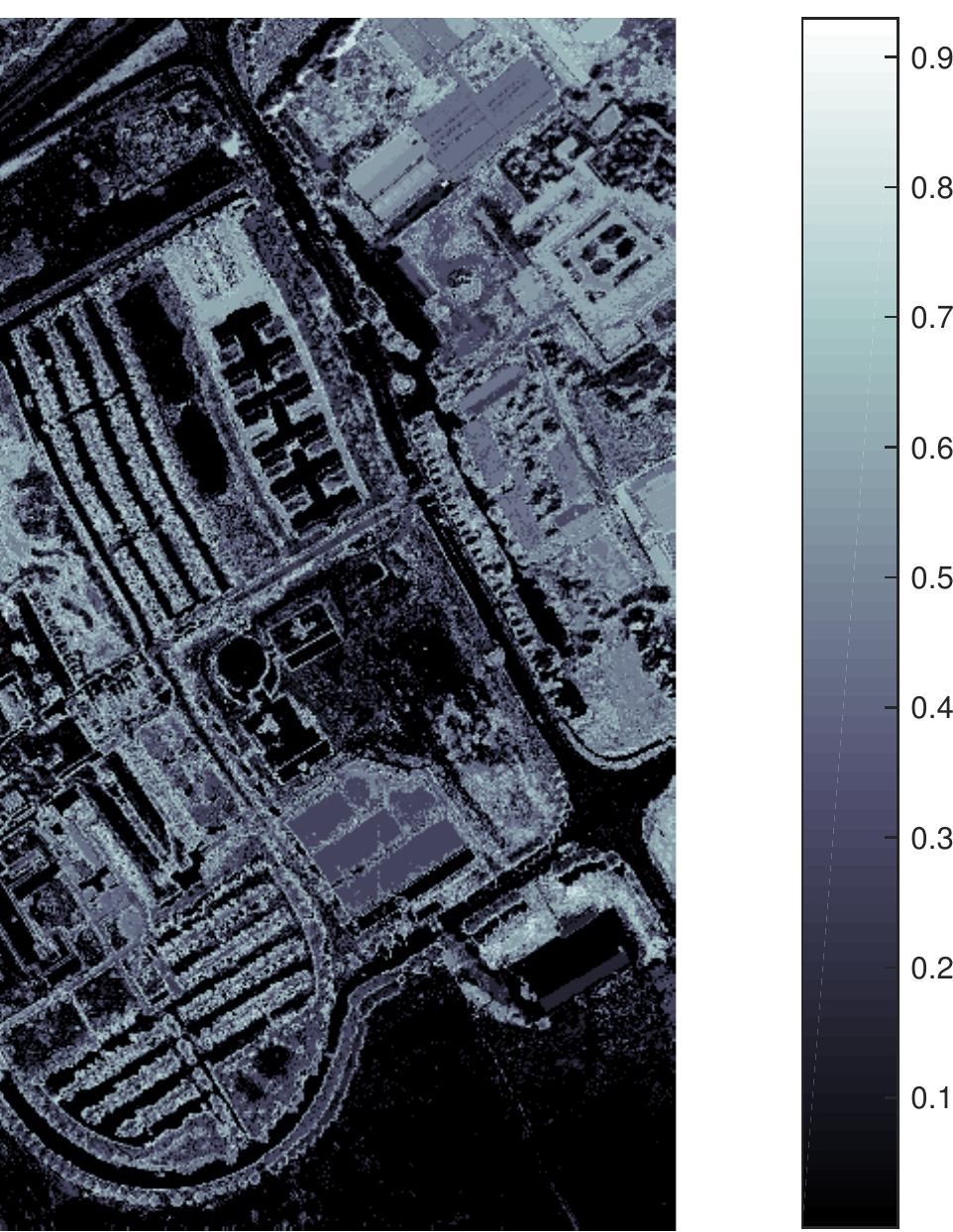}
    \caption{\shiftcbarPaviaU}
    \label{fig:PaviaU_noembedding_pc2_clutter}
  \end{subfigure}
  \hfill
  \begin{subfigure}[b]{\fignocbar}
    \centering
    \includegraphics[height=\heightPaviaU]{%
      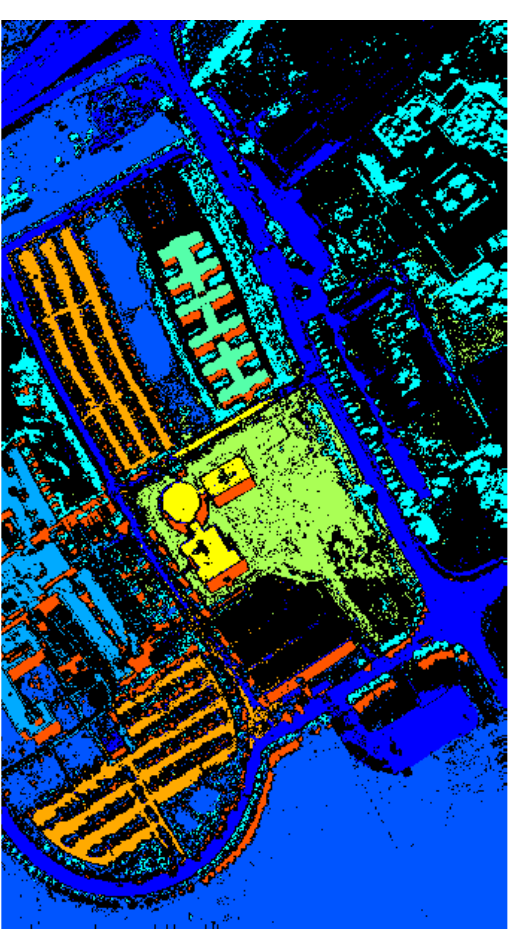}
    \caption{ }
    \label{fig:PaviaU_noembedding_pc2_label_clutter_masked}
  \end{subfigure}
  \caption{%
    Classification and clutter detection results for the University of
    Pavia scene without feature embedding. 
    \emph{(a) and (e)}~RGB composite~\cite{kang_spectral-spatial_2014} and
    manual classification labeling and mask.
    \emph{(b)--(d)}~5\% labeled data sampling: masked classification;
    classification entropy map; classification and clutter removal with
    $\thresClutter = 0.15$.
    \emph{(f)--(h)}~2\% labeled data sampling: same as (b)--(d) with
    $\thresClutter = 0.15$.}
  \label{fig:results-pavia-attributes}
\end{figure*}

In particular, we carry out a set of parallel experiments to those of
\ref{sec:experiments} without application of the spatial-spectral filters;
the ensemble size is consequently reduced to $9$.
Results for the two datasets are shown in
\cref{fig:results-indianpines-attributes,fig:results-pavia-attributes},
respectively; these are analogous to \cref{fig:IndianPines_traditional,%
  fig:PaviaU_traditional}. \Cref{tab:classification-accuracy-attributes}
summarizes the attained classification accuracy, same as
\cref{tab:classification-accuracy-attributes}.
Evidently, the experiments with feature embedding yield higher
classification accuracy, as well as sharper clutter maps and labeled region
boundaries.

We remark also on the improvement extent that may be gained by feature
space embedding. From the class legends provided in
\cref{sec:sup-reference-data}, one may expect a significant difference
between the two datasets, with regard to inter-class similarities.
Indeed, spectral signatures in the India Pines scene are very similar
between certain classes (such as different corn fields, soybean areas, or
grass patches), whereas classes in the University of Pavia scene feature
more distinctive signatures in comparison.
This difference between the datasets means that the former presents a
greater challenge to conventional discrimination metrics, and thereby
benefits more from feature space embedding, which effectively amounts to an
adaptive transformation of the distance metric in the original feature
space. Such benefits are confirmed by our experimental results.

%

%

%
\begin{table*}[t]
  \centering
  \caption{Classification accuracy for the embedding ensemble and
    instances without feature embedding.}
  \label{tab:classification-accuracy-attributes}
  \begin{tabular}{%
    c  
    c<{\%}
    r@{.}l @{$\,\pm\,$} r@{.}l @{$\;\,$} >{(}r@{.}l<{)}  
    r@{.}l @{$\,\pm\,$} r@{.}l @{$\;\,$} >{(}r@{.}l<{)}  
    >{\hshiftacc}r@{.}l  
    >{\hshiftacc}r@{.}l
    }
    \toprule
      \multirow{2}{*}[\vshiftmrow]{\textbf{Dataset}}
    & \multicolumn{1}{c}{%
      \multirow{2}{*}[\vshiftmrow]{\begin{tabular}{@{}c@{}}
                                     \textbf{Labeled}\\[\vshiftmrow]
                                     \textbf{Samples}
                                   \end{tabular}}}
    & \multicolumn{12}{c}{\textbf{Instances} [mean $\pm$ std (max)]}
    & \multicolumn{4}{c}{\textbf{Ensemble}}
    \\
    \cmidrule(lr){3-14}
    \cmidrule(lr){15-18}
    & \multicolumn{1}{c}{}
    & \multicolumn{6}{c}{\textbf{OA} (\%)}
    & \multicolumn{6}{c}{\textbf{AA} (\%)}
    & \multicolumn{2}{c}{\textbf{OA} (\%)}
    & \multicolumn{2}{c}{\textbf{AA} (\%)}
    \\
    \midrule
      \multirow{2}{*}{\begin{tabular}{@{}c@{}}
                        Indian\\[\vshiftmrow]Pines
                      \end{tabular}}
    & 5
    & 76&30   &   3&11   &   73&31
    & 74&37   &   2&66   &   77&42
    & 82&09
    & 78&67
    \\
    & 10
    & 80&18   &   3&09   &   83&14
    & 79&63   &   2&66   &   83&00
    & 85&83
    & 83&77
    \\
    \midrule[0.1pt]
    \multirow{2}{*}{\begin{tabular}{@{}c@{}}
                      University\\[\vshiftmrow]of Pavia
                    \end{tabular}}
    & 2
    & 95&18   &   1&55   &   96&86
    & 94&44   &   1&78   &   96&27
    & 97&53
    & 96&89
    \\
    & 5
    & 96&82   &   1&17   &   97&84
    & 96&11   &   1&29   &   97&27
    & 98&60
    & 98&13
    \\
    \bottomrule
  \end{tabular}
\end{table*}



\phantomsection
\label{sec:references}
\addcontentsline{toc}{section}{References}

\ifdefined\localcompile

  \bibliographystyle{ieee}
  \bibliography{../../../refs/references}

\else

\fi

\end{document}